\documentclass[twoside]{article}

\usepackage[utf8]{inputenc}  
\usepackage[T1]{fontenc}     

\usepackage[authoryear]{natbib}

\usepackage{microtype}       
\usepackage{nicefrac}        
\usepackage{url}      
\usepackage{xurl}     
\usepackage[misc]{ifsym}

\usepackage{amsmath}         
\usepackage{amssymb}         
\usepackage{amsthm}          
\usepackage{amsfonts}        

\usepackage{booktabs}        
\usepackage{array}           
\usepackage{multirow}        
\usepackage{adjustbox}       
\usepackage[table]{xcolor}  
\usepackage{colortbl}       
\definecolor{FairSHAPBlue}{RGB}{220,235,255}
\newcommand{\FSB}[1]{\cellcolor{FairSHAPBlue}#1}
\usepackage{enumitem}        
\usepackage{tikz}            

\usepackage{algorithm}       
\usepackage{algorithmicx}    
\usepackage[noend]{algpseudocode} 
\newcommand{\LineComment}[1]{\hfill \textcolor{blue!20!black}{\(//\) #1}}
\newcommand{\topelement}[1]{\footnotesize{$\pm$#1}}

\usepackage{titlesec}        
\usepackage{titletoc}        
\usepackage{tocloft}         
\definecolor{darkblue}{rgb}{0.0, 0.0, 0.55}
\usepackage[
    colorlinks=true,         
    linkcolor=darkblue,         
    citecolor=darkblue,         
    urlcolor=darkblue           
]{hyperref}

\usepackage{appendix}        
\let\appendixpagenameorig\appendixpagename
\renewcommand{\appendixpagename}{\normalfont\LARGE\scshape\appendixpagenameorig}

\usepackage[acronym]{glossaries}
\glsdisablehyper
\newacronym{dr}{DR}{discriminative risk}
\newacronym{dp}{DP}{demographic parity}
\newacronym{eo}{EO}{equality of opportunity}
\newacronym{pqp}{PQP}{predictive quality parity}

\theoremstyle{plain}
\newtheorem{theorem}{Theorem}[section]
\newtheorem{proposition}[theorem]{Proposition}
\newtheorem{lemma}[theorem]{Lemma}

\theoremstyle{definition}
\newtheorem{definition}[theorem]{Definition}

\theoremstyle{remark}

\def\myDR{\mathrm{DR}}

\AtBeginDocument{%
    \setlength\abovedisplayskip{3pt}%
    \setlength\belowdisplayskip{3pt}%
}

\makeatletter
\let\oldmakefirstuc\makefirstuc
\renewcommand*{\makefirstuc}[1]{%
  \def\gls@add@space{}%
  \mfu@capitalisewords#1 \@nil\mfu@endcap
}
\def\mfu@capitalisewords#1 #2\mfu@endcap{%
  \def\mfu@cap@first{#1}%
  \def\mfu@cap@second{#2}%
  \gls@add@space
  \oldmakefirstuc{#1}%
  \def\gls@add@space{ }%
  \ifx\mfu@cap@second\@nnil
    \let\next@mfu@cap\mfu@noop
  \else
    \let\next@mfu@cap\mfu@capitalisewords
  \fi
  \next@mfu@cap#2\mfu@endcap
}
\makeatother

\usepackage[accepted]{aistats2026}
\begin{document}
%

%

\twocolumn[
\aistatstitle{FairSHAP: Preprocessing for Fairness Through Attribution-Based Data Augmentation}

\aistatsauthor{
  Lin Zhu\footnotemark[1] \And
  Yijun Bian\footnotemark[1] \And
  Lei You \textnormal{\textsuperscript\Letter}
}

\aistatsaddress{
  Technical University of Denmark \And
  University of Copenhagen \And
  Technical University of Denmark
}
]

\begin{abstract}
Ensuring fairness in machine learning models is critical, particularly in high-stakes domains where biased decisions can lead to serious societal consequences. However, existing preprocessing approaches generally lack transparent mechanisms for identifying which features are responsible for unfairness. This obscures the rationale behind data modifications.
We introduce FairSHAP, a novel preprocessing framework that leverages Shapley value attribution to improve both individual and group fairness. FairSHAP identifies fairness-critical features in the training data using an interpretable measure of feature importance, and systematically modifies them through instance-level matching across sensitive groups. Our method effectively reduces discriminative risk (DR) with an instance-wise guarantee up to an interaction residual term, which is bounded under local matching, while simultaneously bounding the upper limit of demographic parity (DP), which in practice leads to its reduction. Experiments on multiple tabular datasets show that we achieve state-of-the-art or comparable performance across DR, DP, and equality of opportunity (EO) with minimal modifications, thereby preserving data fidelity. As a model-agnostic and transparent method, FairSHAP integrates seamlessly into existing machine learning pipelines and provides actionable insights into the sources of bias. Our code is available on \url{https://github.com/ZhuMuMu0216/FairSHAP}.
\end{abstract}

\section{Introduction}
Machine learning (ML) has become one of the most prominent and rapidly evolving fields in artificial intelligence (AI), with researchers continuously developing methods to improve model accuracy and reliability. Yet in sensitive domains such as healthcare, finance, and law, prioritizing accuracy alone may not serve society’s best interests; fairness and interpretability are equally critical. Well-documented cases, including racial bias in criminal risk assessment tools and gender bias in automated hiring systems~\citep{propublica_compas_analysis,propublica_compas_analysis_re,dastin2018amazon}, highlight how ML models can perpetuate inequities and erode trust. These examples underscore the need for responsible AI approaches that balance predictive performance with fairness and transparency.

To address bias in ML models, mitigation methods are often categorized by their intervention stage—pre-processing, in-processing, and post-processing~\citep{mehrabi2021survey,binns2018fairness, 10.1145/3616865}. \textbf{Pre-processing} modifies training data before model learning~\citep{hort2024bias}. Typical strategies include perturbation of features or distributions~\citep{Weerts_Fairlearn_Assessing_and_2023,wang2019repairing,calmon2017optimized,feldman2015certifying,fairlearn_correlation}, reweighting of group instances~\citep{calders2013unbiased,kamiran2012data,arnaiz2023towards}, fair representation learning through invariant embeddings~\citep{pmlr-v28-zemel13,louizos2015variational}, and resampling to balance group distributions~\citep{shekhar2021adaptive,ustun2019fairness,bastani2019probabilistic}. While widely applicable, these methods risk overfitting, data loss, or reduced fidelity when perturbations are excessive.
\textbf{In-processing} integrates fairness as constraints into training. Approaches include adversarial debiasing~\citep{grari2023adversarial,zhang2018mitigating,madras2018learning}, constrained optimization~\citep{zafar2017fairness,agarwal2018reductions}, fairness-aware regularization~\citep{kamishima2012fairness,donini2018empirical,zhao2022towards}, reweighted gradient updates~\citep{wan2023processing}, and counterfactual fairness~\citep{kusner2017counterfactual,chen2025counterfactual}.
\textbf{Post-processing} adjusts model outputs after training. Common methods include threshold tuning for equalized odds~\citep{hardt2016equality}, probability calibration~\citep{dwork2018decoupled}, and boundary modification in uncertain regions~\citep{petersen2021post}. Among these, preprocessing methods are often model-agnostic and widely applicable, but it must carefully balance fairness gains against preserving data integrity and predictive fidelity~\citep{mehrabi2021survey,Weerts_Fairlearn_Assessing_and_2023}.


Despite extensive work on fairness in ML, two key gaps remain. First, fairness and interpretability are often studied in isolation, even though interpretability is essential for uncovering hidden biases and enabling transparent interventions; without their integration, explanations risk being misleading or even “fairwashed\footnote{The term “fairwashing” was popularized by Aïvodji et al. (2019) to describe the risk of rationalizing biased decisions through deceptive explanations.}”~\citep{doshi2017towards, poursabzi2021manipulating, lakkaraju2017identifying, aivodji2019fairwashing}. Second, most methods focus narrowly on group fairness metrics such as demographic parity or equalized odds~\citep{mehrabi2021survey}, thereby overlooking intersectional biases~\citep{kearns2018preventing}, the incompatibility of fairness definitions~\citep{kleinberg2018inherent}, and the importance of individual-level fairness~\citep{dwork2012fairness}.\looseness=-1

Building on these limitations, recent work shows that optimizing only for group fairness can lead to individual arbitrariness~\citep{long2023individual}. Prior research further suggests that individual and group fairness, rather than being contradictory, are different manifestations of the same underlying principles of consistency and egalitarianism~\citep{binns2020apparent}. Under certain conditions, ensuring fairness for individuals can also promote fairness across groups. We therefore argue that strengthening individual fairness provides a principled path toward achieving both individual- and group-level fairness. \textit{Our approach aims to improve both individual and group fairness}.\looseness=-1

Motivated by this insight, we propose an innovative approach that explicitly emphasizes improving individual fairness, hypothesizing that addressing individual-level biases can concurrently enhance group-level fairness. \textit{Our approach positions fairness enhancement as an interpretability-driven task}, where transparency in identifying bias at the individual level facilitates targeted and principled mitigation.
To operationalize this concept, we utilize Shapley values, a widely recognized interpretability metric that quantifies feature importance in individual predictions. Shapley values allow us to pinpoint precisely which features and data points contribute most significantly to fairness disparities. Through targeted adjustments informed by Shapley values, our method ensures data integrity and interpretability while effectively reducing bias.
Our main contributions are as follows:
\begin{itemize}[leftmargin=2em,itemsep=1pt,parsep=1pt,topsep=2pt]
    \item We propose {FairSHAP}, a novel fairness enhancement method leveraging Shapley value analysis and instances matching. It systematically identifies fairness-sensitive data points using individual fairness measures and Shapley values, selectively modifying these points to enhance fairness.
    \item FairSHAP consistently achieves superior fairness improvements with minimal data perturbation, surpassing existing benchmark methods, without requiring extra transformations on the test set.
    \item FairSHAP effectively reduces both individual and group fairness measures (DR, DP, EO) across diverse datasets, with negligible accuracy loss and occasional improvements in predictive performance.
    \item As a preprocessing technique, FairSHAP integrates seamlessly into existing ML workflows, providing fully reproducible and transparent implementations.
\end{itemize}

By aligning fairness enhancement closely with interpretability, our proposed approach paves the way for more accountable, transparent, and equitable ML systems.

\section{Preliminaries}

We summarize some necessarily preliminary knowledge here before elaborate on our methodology.

\subsection{Shapley value and relevant methods}
The Shapley value is a foundational concept in cooperative game theory, originally introduced to fairly distribute a collective payoff among players based on their individual contributions. Given its axiomatic fairness properties, the Shapley value has been widely adopted in various domains, including ML explainability, resource allocation, and feature attribution ~\citep{sundararajan2020many}.

For a set of players (or elements) $\mathcal{N} = \{1, 2, \dots, n\}$ and a characteristic function $v: 2^{\mathcal{N}} \to \mathbb{R}$ that assigns a value to each coalition (or combination) $S \subseteq \mathcal{N}$, the Shapley value of player $k$ is defined as:
\begin{equation}
\phi_k(v) = 
\sum_{S \subseteq \mathcal{N} \setminus \{k\}}
\! \frac{|S|!(n - |S| - 1)!}{n!} 
\big(v(S \cup \{k\}) - v(S)\big)
\,.\label{eq1}
\end{equation}
This formula ensures a fair allocation by averaging the marginal contribution of each player over all possible coalition formations. The key properties of the Shapley value include \textit{efficiency}, \textit{symmetry}, \textit{dummy player}, and \textit{additivity}, making it a unique solution satisfying these fairness criteria.


\textbf{Reference data} \quad In SHAP (SHapley Additive exPlanations) library ~\citep{lundberg2017unified}, reference data serves as a baseline distribution used to simulate the absence of features when computing Shapley values. By replacing features of the instance to be explained with values sampled from the reference data, SHAP estimates the marginal contribution of each feature. The selection of reference data significantly influences the interpretation results.

\textbf{Baseline Shapley} \quad This method is proposed to address the issue that feature values cannot be empty when computing Shapley values using ML~\citep{lundberg2017unified, merrick2020explanation, you2024refining}. It relies on a specific reference data point \( \mathbf{r}_j \) for any instance \( \mathbf{x}_i \), where \( \mathbf{x}_i \) denotes the $i$-th instance, and the set function is defined accordingly:
\begin{equation}
    v_{\text{B}}^{(i,j)}(S) = f(\mathbf{x}_{i,S}; \mathbf{r}_{j,\mathcal{F} \setminus S})-f(\mathbf{r}_j) \,,
\end{equation}
where a feature’s absence is modeled using its value in the reference baseline data point \( \mathbf{r} \), and $\mathcal{F}$ denotes the set of all features. 

\textbf{Random baseline Shapley} \quad This method is proposed to address the issue of feature absence in Shapley value computation by using a random sampling approach from a reference distribution $\mathcal{D}$ ~\citep{lundberg2017unified}, which is a variant of Baseline Shapley and is applied in the famous SHAP library. 
Instead of relying on a single reference data point, it uses a reference distribution to model missing feature values. The set function is defined as follows:
\begin{equation}
    v_{\text{RB}}^{(i)}(S) = \mathbb{E}_{\mathbf{x}' \sim \mathcal{D}} \big[ f(\mathbf{x}_{i,S}; \mathbf{x}_{\mathcal{F} \setminus S}') \big] - \mathbb{E}_{\mathbf{x}' \sim \mathcal{D}} \left[ f(\mathbf{x}') \right]
    \,.
\end{equation}



\subsection{\Gls{dr}, an individual fairness measure}
Since Shapley value can only be applied to individual instances for feature attribution based on the output results, we consider combining it with individual fairness measures rather than group fairness measures. This approach is more suitable for analyzing single instances and enhances the focus on and evaluation of individual fairness, while also providing a means to further mitigate potential biases inherent in group fairness measures.

This approach allows us to leverage both the original data and background data to select feature values that can effectively reduce biases in the training set. In previous work, \citet{bian2023increasing} proposed a critical individual fairness measure named \emph{discriminative risk (\gls{dr})}, and we show it 
(denoted as $L_\text{DR}$) as follows: 
\begin{equation}
\label{equation_dr}
L_\myDR(f,{\mathbf{x}})
= \big|\, f(\check{\mathbf{x}},A=0) - f(\check{\mathbf{x}},A=1) \,\big|
\le \varepsilon \,.
\end{equation}
Here, let $\mathcal{F}$ denote the set of all features. 
An instance is written as $\mathbf{x} = (\check{\mathbf{x}}, A)$, 
where $\check{\mathbf{x}} \in \mathcal{X}_{\mathcal{F}\setminus\{A\}}$ represents the non-sensitive features and $A \in \{0,1\}$ is the sensitive attribute, with \textbf{$0$ corresponding to the unprivileged group(s) and $1$ to the privileged group}. 
A higher $L_{\myDR}(f,\mathbf{x})$ indicates that the model's predictions are overly sensitive to changes in the sensitive attribute, potentially reflecting bias.

For definitions of other group fairness measures, such as \gls{dp}, \gls{eo} and \gls{pqp}, please refer to Appendix~\ref{definition_group_fairness}. The group fairness measures discussed in this paper quantify the disparities between groups. A higher value indicates a greater degree of unfairness; therefore, our objective is to minimize these measures.

\section{The proposed \textbf{FairSHAP} and its theoretical foundation}
\label{headings}
Bias in machine learning models typically stems from two sources: data bias and algorithmic bias. Instead of addressing them separately, we propose tailoring preprocessing strategies to each algorithm on the same dataset, thereby jointly mitigating both sources of bias. This approach is novel in that it adapts preprocessing to algorithm-specific characteristics while remaining independent of model internals—model-agnostic at the architectural level, but algorithm-adaptive in design. In doing so, it retains broad applicability while offering a more principled way to integrate data and algorithmic considerations in fairness enhancement.


\subsection{Framework of FairSHAP}
To address this problem, we propose FairSHAP, a pre-processing method that enhances individual fairness by identifying fairness-critical positions in the training set using Shapley values, followed by targeted data augmentation. The method aims to reduce the individual fairness measure \gls{dr} as a primary objective, while also leading to reductions in group fairness measures such as \gls{dp} and \gls{eo}. 
Unlike generic data preprocessing approaches that operate without any model information, our method consults the model once at the beginning to obtain prior knowledge about fairness-critical features. Importantly, this single interaction does not modify the model itself; it only guides the preprocessing step to produce a minimally modified dataset. After this preprocessing stage, the model will be subsequently trained on adjusted new dataset, ensuring that FairSHAP remains a data-level intervention rather than a model-level modification. Therefore, we regard FairSHAP as belonging to the preprocessing approaches.\footnote{Feature engineering is standard preprocessing in traditional ML, even when guided by model feedback. Thus, using model information at the preprocessing stage does not contradict the notion of preprocessing.}

\begin{figure*}[tbh]
    \centering
    \includegraphics[width=.981\linewidth]{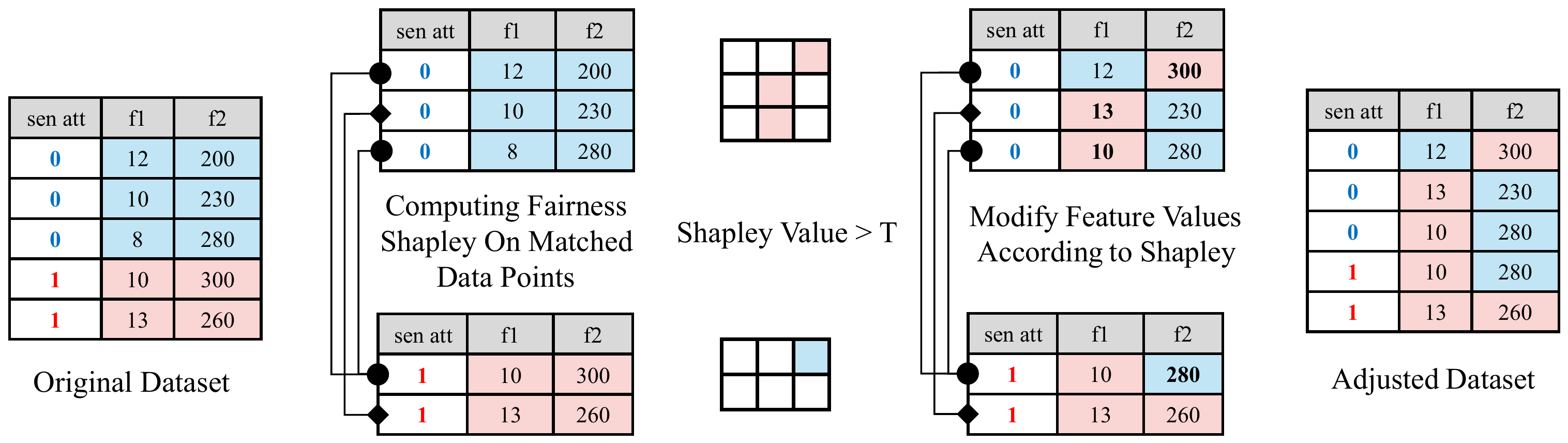}
    \caption{Overall framework of FairSHAP. (Left) Training data are first split by sensitive attribute and aligned via nearest-neighbor matching to produce paired instances. (Right) For each target group, feature values whose Shapley  value exceeds a threshold are adjusted to reduce \gls{dr}, and the modified instances from both groups are recombined into an augmented, fairness-improved training set.}
    \label{fig:overall}
\end{figure*}
FairSHAP consists of two interconnected components, which is presented in Figure~\ref{fig:overall}. The left-hand side of the illustration corresponds to the first component, while the right-hand side depicts the second component. These components work together to identify and mitigate bias in the training data. Below, we describe each component in detail:

\textit{(1)} On the left side of the figure, the process begins by dividing the training set into two groups (privileged group and unprivileged group) based on the sensitive attribute (e.g., gender, race, etc.). To ensure a precise mapping relationship 
between the two groups, a nearest neighbor matching procedure is applied. This step aligns instances from one group with similar instances in the other group, creating matched pairs that facilitate a more equitable analysis. This alignment is visually emphasized in the left portion of the figure, highlighting the pairing process.

\textit{(2)} On the right side of the figure, the focus shifts to leveraging these matched pairs for bias mitigation. Specifically, one group is selected as the target group, while the other group serves as the reference data for computing Shapley values for instances within the target group. These Shapley values are combined with \gls{dr} to identify specific feature values that contribute to fairness disparities. Once identified, these feature values are modified to reduce \gls{dr}. The process is repeated with the roles of the two groups reversed, ensuring that each group can effectively learn useful feature values from the other. This bidirectional learning approach helps to capture a more balanced and comprehensive representation of the data, ultimately improving fairness and reducing disparities. Finally, the modified instances from both groups are combined to form a new  augmented training set, which is designed to reduce biases and improve fairness.

\begin{algorithm}[t]
\caption{Overall framework: Enhancing fairness via matching and Shapley values}
\label{alg:overall_framework}
\begin{algorithmic}[1]
\Require Model $f$, dataset $\mathcal{D} \in \mathbb{R}^{(n+m)\times d}$ with sensitive attribute $A\in\{0,1\}$, threshold $T$, matching method $\mathcal{M}_{\text{method}}$, where $\text{method}\in\{\text{NearestNeighbor},\text{OptimalTransport}\}$
\Ensure $\mathcal{D}_{\text{new}}$

\State Split $\mathcal{D}$ into two subgroups: 
$\mathcal{G}\in\mathbb{R}^{n\times d}$ (e.g., $A=0$) and 
$\widetilde{\mathcal{G}}\in\mathbb{R}^{m\times d}$ (e.g., $A=1$)

\State $\mathcal{G}' \gets \texttt{FairSHAP}(\text{target}=\mathcal{G},\;\text{non-target}=\widetilde{\mathcal{G}},\;\text{model}=f,\;T,\;\mathcal{M}_{\text{method}})$ \LineComment{see Algorithm~\ref{alg:FairSHAP}}
\State $\widetilde{\mathcal{G}}' \gets \texttt{FairSHAP}(\text{target}=\widetilde{\mathcal{G}},\;\text{non-target}=\mathcal{G},\;\text{model}=f,\;T,\;\mathcal{M}_{\text{method}})$
\LineComment{see Algorithm~\ref{alg:FairSHAP}}

\State $\mathcal{D}_{\text{new}} \gets \text{Concat}(\mathcal{G}',\,\widetilde{\mathcal{G}}') \in \mathbb{R}^{(n+m)\times d}$
\State \Return $\mathcal{D}_{\text{new}}$
\end{algorithmic}
\end{algorithm}
\paragraph{Overall framework:}  
The detailed overall process is described in the following algorithm~\ref{alg:overall_framework}.  
Given a dataset $\mathcal{D} \in \mathbb{R}^{(n + m) \times d}$, we partition it into two disjoint groups (privileged group and unprivileged group), $\mathcal{G}\in \mathbb{R}^{n \times d}$ and $\widetilde{\mathcal{G}}\in \mathbb{R}^{m \times d}$, based on the sensitive attribute $A \in \{0, 1\}$, where $0$ and $1$ represent the unprivileged and privileged groups, respectively. Note that the unprivileged group(s) in the original dataset may possibly include multiple values. Subsequently, we treat $\mathcal{G}$ and $\widetilde{\mathcal{G}}$ as the target group in turn and input them into FairSHAP (see algorithm~\ref{alg:FairSHAP}), obtaining the modified groups $\mathcal{G}'$ and $\widetilde{\mathcal{G}}'$, which are then concatenated to form $\mathcal{D}_{\text{new}}$ for training a new model. 
In this process, we also set the threshold $T$ that determines which Shapley values trigger modifications. We set $T{=}0.05$ as a default threshold balancing fairness improvement and data fidelity.
Empirically, modification counts concentrate within $\phi \in[0.01,0.05]$ and show diminishing returns beyond this range, while too small/large thresholds respectively introduce many near-zero edits or miss fairness-critical features (Appendix~\ref{app:threshold}, Fig.~\ref{fig:threshold_counts}).
We observe stable behavior for $T\in[0.03,0.07]$ and thus use $T{=}0.05$ for all experiments.

Next, as shown in Algorithm~\ref{alg:FairSHAP}, we denote by $\mathbf{g}$ and $\tilde{\mathbf{g}}$ the random variables drawn from $\mathcal{G}$ and $\widetilde{\mathcal{G}}$, respectively. For each $\mathbf{g}\in\mathcal{G}$, we identify its closest counterpart $\tilde{\mathbf{g}}\in\widetilde{\mathcal{G}}$, and normalize this matching so that it forms a joint distribution $\mathcal{P}\in\mathbb{R}^{n\times m}$ over the two groups. Importantly, when nearest-neighbor matching is conducted, we not only align instances across sensitive groups (e.g., male versus female), but also enforce label consistency, such that positive-labeled instances in one group are matched exclusively with positive-labeled instances in the other, and analogously for negative-labeled cases. Based on the matched pairs, we then compute the Shapley values $\phi_k$, which quantify the average marginal contribution of the $k$-th feature in an instance to the model prediction relative to a baseline. Features whose Shapley values exceed a threshold $T$ are regarded as fairness-critical and replaced with values from their matched references. This procedure is applied symmetrically to both groups, thereby ensuring balanced fairness improvements.
\begin{algorithm}[t]
\caption{\texttt{FairSHAP}}
\label{alg:FairSHAP}
\begin{algorithmic}[1]
  \Require Target group $\mathcal{G}\in \mathbb{R}^{n\times d}$ and non-target group  $\widetilde{\mathcal{G}}\in \mathbb{R}^{m\times d}$, model $f$, threshold $T$, matching method $\mathcal{M}_{\text{method}}$
  \Ensure modified dataset $\mathcal{G}'$
  \State Use $\mathcal{M}_{\text{method}}(\mathcal{G},\widetilde{\mathcal{G}})$ to obtain joint probability $\mathcal{P}(\mathbf{g},\tilde{\mathbf{g}})\in \mathbb{R}^{n\times m}$

  \State Use Eqs.~\eqref{eq1} and~\eqref{eq5} to obtain Shapley value matrix $\boldsymbol{\phi}\in \mathbb{R}^{n\times d}$ 

  \State Initialize reference data 
    $\mathcal{B}\leftarrow \mathbf{0}_{n\times d}$

  \For{$i = 1$ \textbf{to} $n$}
    \State $j^* \leftarrow \arg\max_{1 \le j \le m} \mathcal{P}_{i,j}$
    \State $\mathcal{B}_{i,:} \leftarrow \widetilde{\mathcal{G}}_{j^*,:}$
  \EndFor

  \State Let $\mathcal{G}' \leftarrow \mathcal{G} (\mathcal{G}' \in \mathbb{R}^{n\times d})$
  \For{$i = 1$ \textbf{to} $n$}
    \For{$k = 1$ \textbf{to} $d$}
      \If{$\phi_{i,k} \ge T$}
        \State $\mathcal{G}'_{i,k} \leftarrow \mathcal{B}_{i,k}$
      \EndIf
    \EndFor
  \EndFor

  \State \Return $\mathcal{G}'$
\end{algorithmic}
\end{algorithm}

Finally, the modified groups $\mathcal{G}'$ and $\widetilde{\mathcal{G}}'$ are combined to create a new dataset $\mathcal{D}_{\text{new}}$, which is used for retraining the model with improved fairness.
Note that
\begin{subequations}
\begin{align}
v^{(i)}(S)
=&\quad \mathbb{E}_{\tilde{\mathbf{g}} \sim \mathcal{P}(\tilde{\mathbf{g}} \mid \mathbf{g}_i)} 
\left[ L_{\myDR}(\mathbf{g}_{i, S}; \tilde{\mathbf{g}}_{\mathcal{F} \setminus S}) \right] \nonumber\\
& - \mathbb{E}_{\tilde{\mathbf{g}} \sim \mathcal{P}(\tilde{\mathbf{g}} \mid \mathbf{g}_i)} 
\left[ L_{\myDR}(\tilde{\mathbf{g}}_{\mathcal{F}}) \right] 
\label{eq:shap_pair_a} \,,\\
\text{s.t.} &\quad 
    \mathcal{P}(\mathbf{g}, \tilde{\mathbf{g}}) 
    = \mathcal{M}_{\text{method}}(\mathcal{G}, \widetilde{\mathcal{G}}) \,,
    \label{eq:shap_pair_b}
\end{align}%
\label{eq5}%
\end{subequations}%
%
%
where \( L_{\myDR} \) 
is the individual fairness measure, with \gls{dr} being used as individual fairness in this paper, \( S \) represents a subset of features considered in the fairness evaluation, and \( \mathcal{F} \) denotes the full set of features, with \( \mathcal{F} \setminus S \) indicating the complement of \( S \). The matching method can be instantiated as nearest neighbor, optimal transport, or any other appropriate similarity-based matching approach.

%

\subsection{Theoretical foundation}

We provide the theoretical guarantees of FairSHAP in reducing \gls{dr}.  
The key idea is that features with large Shapley contributions are fairness-critical; replacing them with values from reference data can improve individual fairness.

\begin{theorem}[Instance‐wise DR reduction]
Fix a threshold $T > 0$ and define the set of fairness-critical features as
\[
S_T({\mathbf{x}}) = \{\, k \in [d] \;\mid\; \phi_k({\mathbf{x}}) \ge T \,\} 
\,,
\]
where $[d]$ denotes $\{1,2,...,d\}$ for brevity. 
Let ${\mathbf{x}}_{\mathrm{after}}$ denote the feature vector obtained from ${\mathbf{x}}$ after FairSHAP replaces every feature $k \in S_T({\mathbf{x}})$ by its value in the matched reference ${\mathbf{r}}$.  
Then the \gls{dr} satisfies
\begin{equation}
\label{eq:fairshap_dr}
\begin{aligned}
L_{\myDR}(f, \mathbf{x}_{\mathrm{after}}) \!
&= L_{\myDR}(f, \mathbf{x}_{\mathrm{before}}) 
-\! \sum_{k \in \mathcal{S}_T(\mathbf{x})} \! \phi_k(\mathbf{x}) + \mathcal{I}(\mathbf{x}) \\
&\le L_{\myDR}(f, \mathbf{x}_{\mathrm{before}}) - T \, |\mathcal{S}_T(\mathbf{x})| + \mathcal{I}(\mathbf{x})
\,.
\end{aligned}
\end{equation}
where \(\mathcal{I}({\mathbf{x}})\) denotes the interaction residual term.
\end{theorem}
This theorem guarantees that each replaced feature has a Shapley-assigned contribution of at least \(T\)
toward reducing the instance-wise \gls{dr}, up to the interaction residual \(\mathcal{I}(\mathbf{x})\). The detailed proof is provided in appendix~\ref{thm:dr-reduction}. To rigorously justify that this interaction residual is negligible in practice, we establish the following proposition. It demonstrates that our use of Nearest Neighbor or Optimal Transport for local matching ensures \(\mathcal{I}(\mathbf{x})\) decays as a second-order term, thereby validating the dominance of the Shapley attribution:

\begin{figure*}[h!]
    \centering
    \includegraphics[width=.971\linewidth]{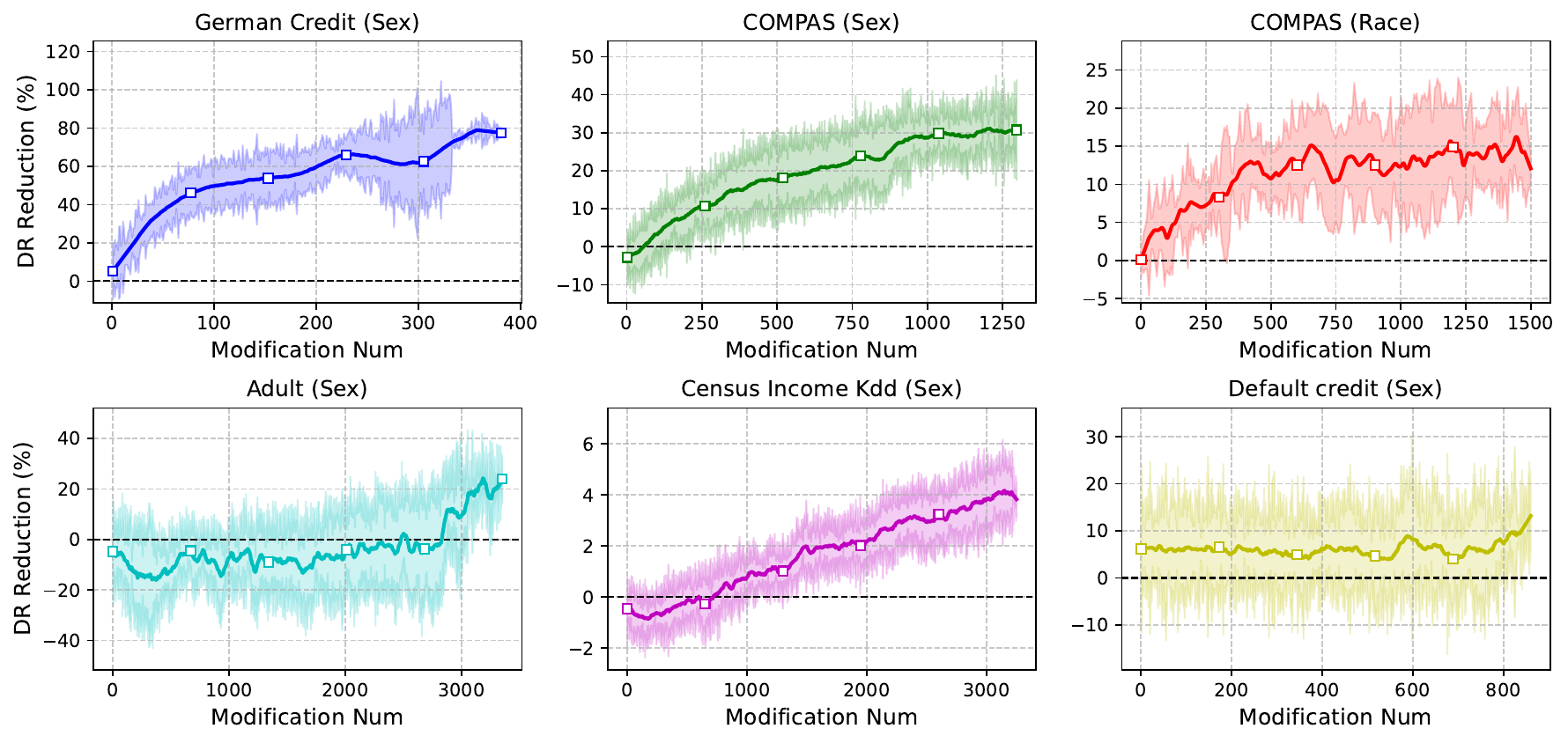}
    \vspace{-1em}
    \caption{Percentage reduction in the \gls{dr} across different datasets.
The $x$-axis denotes the number of modifications applied (up to the maximum required under a fairness threshold $T = 0.05$), while the $y$-axis indicates the relative in \gls{dr}, expressed as a percentage of the original value.}
\label{fig: FairSHAP on DR across all datasets}
\end{figure*}

\begin{proposition}[Bound on Interaction Residual]
Assume the classifier \(f\) is twice differentiable with a bounded Hessian. Since FairSHAP uses Nearest Neighbor or Optimal Transport to select a reference \(\mathbf{r}\) from the local neighborhood of \(\mathbf{x}\), let \(\epsilon = \|\mathbf{x} - \mathbf{r}\|\) be the matching distance. Then, the interaction residual is bounded by a second-order term:
\[
|\mathcal{I}(\mathbf{x})| \le C \cdot \epsilon^2 \,,
\]
where \(C\) is a constant depending on the curvature of \(f\). Consequently, as \(\epsilon \to 0\), the residual \(\mathcal{I}(\mathbf{x})\) vanishes faster than the Shapley attribution \(\phi_k(\mathbf{x})\) (which is \(O(\epsilon)\)), theoretically justifying the approximation in Theorem~\ref{thm:dr-reduction}.
\end{proposition}

Proposition~\ref{prop:residual_bound} relies on the premise that the matching distance \(\epsilon = \|\mathbf{x} - \mathbf{r}\|\) remains sufficiently small to suppress higher-order terms. We empirically validate this assumption in Appendix~\ref{app:matching_distance}. Our analysis of the matching distance distribution confirms that the proposed Nearest Neighbor Matching strategy consistently selects reference samples within the local neighborhood of \(\mathbf{x}\), thereby ensuring the validity of the approximation in practice.
\vspace{-3mm}
\paragraph{Complexity (two-sided).}
Let $n$ and $m$ be the sample sizes of the two sensitive groups, and $d$ the feature dimension.
With nearest neighbor matching, building the dense $n\times m$ cross-group cost matrix costs $\mathcal{O}(nmd)$
time and $\mathcal{O}(nm)$ space. We estimate Shapley values using $M$ permutations and $R$
reference draws per set-function evaluation; let $C_f$ be the cost of one forward pass of $f$.
The total cost is $\mathcal{O}\!\big(nmd + (n{+}m)\,MdRC_f\big)$ time and
$\mathcal{O}\!\big(nm + (n{+}m)d\big)$ space; see Appendix~\ref{app:complexity} for details.

Let $\mathcal{X}$ denote the feature space.  
Each instance is written as $\mathbf{x} = (\check{\mathbf{x}}, A)$,  
where $\check{\mathbf{x}} \in \mathcal{X}_{\mathcal{F}\setminus\{A\}}$ 
denotes the non-sensitive features. 
We denote by $D_a(\check{\mathbf{x}})$ the conditional distribution of non-sensitive features given $A=a\in\{0,1\}$.  

%
\begin{theorem}[Identical‐distribution case]
\label{thm:dp-dr-identical_main}
If \(D_0=D_1\), then for any classifier \(f\),
\[
  L_{\mathrm{DP}}(f) \;\le\; L_{\myDR}(f) \,.
\]
\end{theorem}

\begin{theorem}[General case]
\label{thm:dp-dr-general_main}
For any classifier \(f\), regardless of \(D_0\) and \(D_1\),
\[
\begin{aligned}
  L_{\mathrm{DP}}^{\text{score}}(f) 
  \;\le\;& \hspace{1em} \min\{\,L_\myDR(f;D_0),\,L_\myDR(f;D_1)\,\} \\
  &+ \mathrm{TV}(D_0,D_1) \,.
\end{aligned}
\]
where \(\mathrm{TV}(D_0,D_1)\) is the total variation distance.
\end{theorem}

Theorems~\ref{thm:dp-dr-identical} and~\ref{thm:dp-dr-general} establish that any uniform reduction in individual fairness measure \(L_{\mathrm{DR}}\) induced by FairSHAP also yields a corresponding decrease in \(L_{\mathrm{DP}}\).  Detailed proofs are deferred to Appendix~\ref{app3:dr-dp}.

Thus, our theoretical analysis not only justifies the design of FairSHAP but also highlights its ability to simultaneously improve both individual and group fairness guarantees in a principled and interpretable manner.

\section{Experiments}
We conducted a comprehensive experimental evaluation to assess the proposed method's performance across five critical fairness measures: accuracy, \gls{dr}, \gls{dp}, \gls{eo}, and \gls{pqp}. 

Our experiments incorporates 5 typical tabular datasets spanning criminal justice, finance, and social services domains: {COMPAS}~\citep{propublica_compas_analysis,propublica_compas_analysis_re}, {ADULT}~\citep{adult_2}, {German Credit}~\citep{south_german_credit_573}, {Census Income KDD}~\citep{census_income_kdd_117}, and {Default Credit}~\citep{default_of_credit_card_clients_350}. 


The comparative analysis included a baseline model without bias mitigation mechanisms, and three benchmark preprocessing methods: Remove Sensitive Attribute (RSA), Disparate Impact Remover (DIR)~\citep{feldman2015certifying} and Correlation Remover (CR)~\citep{fairlearn_correlation}. In the following, we refer to them simply as RSA, DIR and CR, respectively.

\begin{table*}[t]
    \centering\vspace{-3mm}
    \caption{Compare FairSHAP with other fairness mitigation methods across different datasets. Here, ``s.a.'' denotes the sensitive attribute used in fairness evaluation, and ``Kdd'' is an abbreviation for the Census Income KDD dataset. Note that TrainingAR: Training Set Adjustment Rate; TestAN: Test Set Adjustment Necessity. Data Fidelity is measured using the Wasserstein Distance to quantify the difference between the original and adjusted data distributions. Best and second-best results for fairness and efficiency metrics (DR, DP, EO, PQP, Data Fidelity, TrainingAR) are \textbf{bold} and \underline{underlined}, respectively.}
    \label{tab:fairness_comparison_extended}
    \vspace{1.5mm}
    \renewcommand\tabcolsep{4.15pt}
    \resizebox{\textwidth}{!}{  
    \begin{tabular}{ll rr rrr rrc}
        \toprule
        \textbf{Dataset (s.a.)} & \textbf{Methods} 
        & \multicolumn{1}{c}{\textbf{Accuracy}} 
        & \multicolumn{1}{c}{\textbf{DR}} 
        & \multicolumn{1}{c}{\textbf{DP}} 
        & \multicolumn{1}{c}{\textbf{EO}} 
        & \multicolumn{1}{c}{\textbf{PQP}} 
        & \multicolumn{1}{c}{\textbf{Data Fidelity}} 
        & \textbf{TrainingAR} 
        & \textbf{TestAN} \\
        \midrule
        \multirow{5}{*}{German (sex)} 
        & Baseline & 0.6650\topelement{0.0257} & 0.0785\topelement{0.0211} & 0.0512\topelement{0.0346} & 0.1287\topelement{0.0590} & 0.1341\topelement{0.0486} & --- & --- & No \\
        & RSA & 0.6590\topelement{0.0287} & --- & 0.0522\topelement{0.0246} & 0.1524\topelement{0.1007} & 0.2189\topelement{0.0815} & --- & --- & Yes \\
        & CR & \underline{0.6680\topelement{0.0238}} & \textbf{0.0028\topelement{0.0029}} & 0.0844\topelement{0.0557} & 0.1559\topelement{0.0609} & \textbf{0.0723\topelement{0.0330}} & 0.0183\topelement{0.0211} & 0.9615 & Yes \\
        & DIR & \textbf{0.6720\topelement{0.0337}} & 0.0966\topelement{0.0112} & 0.0946\topelement{0.0373} & 0.1737\topelement{0.0729} & 0.1529\topelement{0.0634} & 0.0155\topelement{0.0440} & 0.0774 & Yes \\
        & \FSB{FairSHAP} & \FSB{0.6630\topelement{0.0275}} & \FSB{\underline{0.0243\topelement{0.0112}}} & \FSB{\textbf{0.0301\topelement{0.0347}}} & \FSB{\textbf{0.1126\topelement{0.0783}}} & \FSB{0.1852\topelement{0.1074}} & \FSB{\textbf{0.0049\topelement{0.0085}}} & \FSB{\textbf{0.0156}} & \FSB{\textbf{No}} \\
        \midrule
        \multirow{5}{*}{COMPAS (sex)} 
        & Baseline & \textbf{0.6698\topelement{0.0051}} & 0.0883\topelement{0.0064} & 0.1548\topelement{0.0241} & 0.1243\topelement{0.0510} & 0.0492\topelement{0.0084} & --- & --- & No \\
        & RSA & 0.6676\topelement{0.0067} & --- & \textbf{0.1075\topelement{0.0160}} & \textbf{0.0842\topelement{0.0475}} & 0.0635\topelement{0.0266} & --- & --- & Yes \\
        & CR & 0.6679\topelement{0.0045} & \textbf{0.0082\topelement{0.0070}} & 0.1407\topelement{0.0248} & 0.1291\topelement{0.0317} & 0.0714\topelement{0.0517} & 0.0189\topelement{0.0193} & 0.9174 & Yes \\
        & DIR & 0.6644\topelement{0.0098} & 0.1150\topelement{0.0091} & \underline{0.1155\topelement{0.0239}} & 0.0952\topelement{0.0359} & 0.0747\topelement{0.0370} & 0.0387\topelement{0.0640} & 0.0650 & Yes \\
        & \FSB{FairSHAP} & \FSB{0.6609\topelement{0.0106}} & \FSB{\underline{0.0629\topelement{0.0091}}} & \FSB{0.1326\topelement{0.0407}} & \FSB{\underline{0.0985\topelement{0.0603}}} & \FSB{\textbf{0.0452\topelement{0.0383}}} & \FSB{\textbf{0.0025\topelement{0.0048}}} & 
        \FSB{\textbf{0.0113} & \FSB{\textbf{No}}} \\
        \midrule
        \multirow{5}{*}{COMPAS (race)} 
        & Baseline & \textbf{0.6689\topelement{0.0108}} & 0.0995\topelement{0.0076} & 0.1436\topelement{0.0209} & 0.1438\topelement{0.0233} & 0.0522\topelement{0.0406} & --- & --- & No \\
        & RSA & 0.6623\topelement{0.0078} & --- & 0.2069\topelement{0.0128} & 0.2024\topelement{0.0349} & 0.0575\topelement{0.0291} & --- & --- & Yes \\
        & CR & 0.6611\topelement{0.0112} & \textbf{0.0418\topelement{0.0092}} & 0.1502\topelement{0.0341} & 0.1621\topelement{0.0530} & 0.0592\topelement{0.0367} & 0.0250\topelement{0.0222} & 0.8920 & Yes \\
        & DIR & 0.6149\topelement{0.0286} & 0.1185\topelement{0.0181} & \underline{0.1359\topelement{0.1241}} & \textbf{0.1117\topelement{0.0945}} & \textbf{0.0399\topelement{0.0338}} & 0.0512\topelement{0.0736}   & 0.0701 & Yes \\
        & \FSB{FairSHAP} & \FSB{0.6627\topelement{0.0069}} & \FSB{\underline{0.0842\topelement{0.0049}}} & \FSB{\textbf{0.1344\topelement{0.0332}}} & \FSB{0.1568\topelement{0.0343}} & \FSB{\underline{0.0508\topelement{0.0469}}} & \FSB{\textbf{0.0040\topelement{0.0055}}}  & \FSB{\textbf{0.0126}} & \FSB{\textbf{No}} \\
        \midrule
        \multirow{5}{*}{Adult (sex)} 
        & Baseline & \textbf{0.8722\topelement{0.0033}} & 0.0315\topelement{0.0037} & 0.1805\topelement{0.0066} & 0.0735\topelement{0.0275} & 0.0275\topelement{0.0321} & --- & --- & No \\
        & RSA & 0.8721\topelement{0.0024} & --- & 0.1770\topelement{0.0066} & \underline{0.0678\topelement{0.0277}} & \textbf{0.0272\topelement{0.0252}} & --- & --- & Yes \\
        & CR & 0.8706\topelement{0.0029} & \textbf{0.0000\topelement{0.0000}} & 0.1824\topelement{0.0055} & 0.0955\topelement{0.0243} & 0.0278\topelement{0.0173} & 0.0167\topelement{0.0391} & 0.9887 & Yes \\
        & DIR & 0.8550\topelement{0.0067} & 0.0499\topelement{0.0076} & \underline{0.1607\topelement{0.0157}} & 0.0772\topelement{0.0624} & 0.0360\topelement{0.0253} & 0.0046\topelement{0.0417} & 0.0081 & Yes 
        \\
        & \FSB{FairSHAP} & \FSB{0.8692\topelement{0.0046}} & \FSB{\underline{0.0273\topelement{0.0047}}} & \FSB{\textbf{0.1558\topelement{0.0130}}} & \FSB{\textbf{0.0393\topelement{0.0254}}} & \FSB{0.0474\topelement{0.0319}} & \FSB{\textbf{0.0010\topelement{0.0073}}} & \FSB{\textbf{0.0012}} & \FSB{\textbf{No}} \\
        \midrule
        \multirow{5}{*}{Adult (race)} 
        & Baseline & \textbf{0.8721\topelement{0.0033}} & 0.0398\topelement{0.0025} & 0.1034\topelement{0.0110} & 0.0808\topelement{0.0326} & 0.0302\topelement{0.0265} & --- & --- & No \\
        & RSA & 0.8710\topelement{0.0044} & --- & 0.0971\topelement{0.0043} & 0.0864\topelement{0.0347} & 0.0427\topelement{0.0142} & --- & --- & Yes \\
        & CR & 0.8713\topelement{0.0033} & \textbf{0.0000\topelement{0.0000}} & 0.1008\topelement{0.0115} & 0.0983\topelement{0.0389} & 0.0480\topelement{0.0235} & 0.0300\topelement{0.0450}  & 0.9620 & Yes \\
        & DIR & 0.8320\topelement{0.0173} & 0.0740\topelement{0.0209} & \textbf{0.0703\topelement{0.0515}} & 0.0871\topelement{0.0355} & 0.0482\topelement{0.0730} & 0.0252\topelement{0.0480} & 0.0089 & Yes 
        \\
        & \FSB{FairSHAP} & \FSB{0.8720\topelement{0.0023}} & \FSB{\underline{0.0284\topelement{0.0017}}} & \FSB{\underline{0.0851\topelement{0.0155}}} & \FSB{\textbf{0.0287\topelement{0.0277}}} & \FSB{\textbf{0.0259\topelement{0.0318}}} & \FSB{\textbf{0.0030\topelement{0.0084}}} & \FSB{\textbf{0.0014}} & \FSB{\textbf{No}} \\
        \midrule
        \multirow{5}{*}{Kdd (sex)} 
        & Baseline & 0.9377\topelement{0.0026} & 0.0767\topelement{0.0011} & 0.0012\topelement{0.0008} & 0.0012\topelement{0.0008} & 0.0796\topelement{0.0054} & --- & --- & No \\
        & RSA & \underline{0.9380\topelement{0.0026}} & --- & \textbf{0.0008\topelement{0.0004}} & \underline{0.0008\topelement{0.0004}} & 0.0796\topelement{0.0055} & --- & --- & Yes \\
        & CR & 0.9377\topelement{0.0025} & \textbf{0.0000\topelement{0.0000}} & 0.0013\topelement{0.0004} & 0.0013\topelement{0.0003} & 0.0796\topelement{0.0054} & 0.0007\topelement{0.0013} & 0.9790 & Yes \\
        & DIR & 0.9377\topelement{0.0028} & 0.0766\topelement{0.0014} & 0.0012\topelement{0.0010} & 0.0013\topelement{0.0011} & 0.0797\topelement{0.0055} & 0.0001\topelement{0.0007} & 0.0013 & Yes \\
        & \FSB{FairSHAP} & \FSB{\textbf{0.9381\topelement{0.0029}}} & \FSB{\underline{0.0732\topelement{0.0020}}} & \FSB{\underline{0.0008\topelement{0.0007}}} & \FSB{\textbf{0.0006\topelement{0.0008}}} & \FSB{\textbf{0.0794\topelement{0.0060}}} & \FSB{\textbf{0.0000\topelement{0.0000}}} & \FSB{\textbf{0.0003}} & \FSB{\textbf{No}} \\
        \midrule
        \multirow{5}{*}{DefaultCredit (sex)} 
        & Baseline & 0.8141\topelement{0.0059} & 0.0226\topelement{0.0019} & 0.0335\topelement{0.0072} & 0.0348\topelement{0.0211} & 0.0269\topelement{0.0127} & --- & --- & No \\
        & RSA & 0.8144\topelement{0.0059} & --- & \textbf{0.0259\topelement{0.0063}} & \underline{0.0244\topelement{0.0122}} & \textbf{0.0154\topelement{0.0172}} & --- & --- & Yes \\
        & CR & \underline{0.8144\topelement{0.0068}} & \textbf{0.0003\topelement{0.0002}} & 0.0339\topelement{0.0077} & 0.0419\topelement{0.0203} & 0.0363\topelement{0.0203} & 0.0051\topelement{0.0064} & 0.9844 & Yes \\
        & DIR & 0.8138\topelement{0.0051} & 0.0224\topelement{0.0027} & 0.0308\topelement{0.0091} & 0.0334\topelement{0.0244} & \underline{0.0255\topelement{0.0214}} & 0.0023\topelement{0.0056} & 0.0741 & Yes 
        \\
        & \FSB{FairSHAP} & \FSB{\textbf{0.8145\topelement{0.0050}}} & \FSB{\underline{0.0214\topelement{0.0026}}} & \FSB{\underline{0.0306\topelement{0.0054}}} & \FSB{\textbf{0.0216\topelement{0.0164}}} & \FSB{0.0289\topelement{0.0075}} & \FSB{\textbf{0.0001\topelement{0.0003}}} & \FSB{\textbf{0.0004}} & \FSB{\textbf{No}} \\
        \bottomrule
    \end{tabular}
    }
\end{table*}

\paragraph{Overall setup}
We used the aforementioned datasets, where the sensitive attribute (s.a.) was preprocessed according to its type.  
When the s.a.\ is \textit{sex}, it is treated as binary (since the sex attribute in all datasets is binary), and we subsequently refer to it simply as s.a. When the s.a.\ is \textit{race}, it naturally involves multiple classes, corresponding to the multi-class case.  
In all scenarios, for the purpose of fairness analysis, we further group the data into a privileged group and an unprivileged group.  
In addition, categorical features were one-hot encoded and numerical features were standardized.  
For model training, we employed the XGBoost Classifier with default hyperparameter settings throughout the project~\citep{chen2016xgboost}. All experiments were conducted using 5-fold cross-validation and conducted on computer equipped with i5-13500H CPU.


\subsection{Main results}
The impact of FairSHAP on \gls{dr} is shown in Figure~\ref{fig: FairSHAP on DR across all datasets}.
Each curve represents the average over 5 folds, with shaded areas indicating 
standard deviation.
The dashed horizontal line at $y = 0$ corresponds to the original \gls{dr} of the unmodified model.
This figure compares the relative \gls{dr} reduction of a model retrained on a FairSHAP-modified training set and evaluated on the test set, against the original model evaluated on the same test set. Higher percentages indicate a greater reduction in \gls{dr} compared to the original model, and thus reflect improved fairness.

\begin{table*}[t]
    \centering\vspace{-3mm}
    \caption{Compare FairSHAP with ablation studies across different datasets. The sensitive attribute is sex in all cases. Best and second-best results for fairness measures (DR, DP, EO, PQP) are \textbf{bold}.}
    \label{tab:fairness_comparison_ablation}\vspace{1mm}
    \renewcommand\tabcolsep{4.7pt}
    \scalebox{.74}{
    \begin{tabular}{llc rr rrr}
        \toprule
        \textbf{Dataset} & \multicolumn{1}{c}{\textbf{Methods}} 
        & ~ & \multicolumn{1}{c}{\textbf{Accuracy}} 
        & \multicolumn{1}{c}{\textbf{DR}} 
        & \multicolumn{1}{c}{\textbf{DP}} 
        & \multicolumn{1}{c}{\textbf{EO}} 
        & \multicolumn{1}{c}{\textbf{PQP}} \\
        \midrule
        \multirow{4}{*}{German} 
        & Baseline & & 0.6650\topelement{0.0257} & 0.0785\topelement{0.0211} & 0.0512\topelement{0.0346} & 0.1287\topelement{0.0590} & \textbf{0.1341\topelement{0.0486}} \\
        & Ablation study 1 & & \textbf{0.6690\topelement{0.0237}} & 0.0709\topelement{0.0239} & 0.0446\topelement{0.0106} & 0.0972\topelement{0.1091} & 0.1463\topelement{0.1133} \\
        & Ablation study 2 & & 0.6470\topelement{0.0452} & 0.0640\topelement{0.0176} & 0.0708\topelement{0.0510} & 0.1619\topelement{0.0730} & 0.1475\topelement{0.1384} \\
        & FairSHAP & & 0.6630\topelement{0.0275} & \textbf{0.0243\topelement{0.0112}} & \textbf{0.0301\topelement{0.0347}} & \textbf{0.1126\topelement{0.0783}} & 0.1852\topelement{0.1074} \\
        \midrule
        \multirow{4}{*}{COMPAS} 
        & Baseline & & 0.6698\topelement{0.0051} & 0.0883\topelement{0.0064} & 0.1548\topelement{0.0241} & 0.1243\topelement{0.0510} & 0.0492\topelement{0.0084} \\
        & Ablation study 1 & & \textbf{0.6713\topelement{0.0090}} & 0.0903\topelement{0.0110} & 0.1519\topelement{0.0114} & 0.1315\topelement{0.0379} & 0.0490\topelement{0.0465} \\        
        & Ablation study 2 & & 0.6699\topelement{0.0076} & 0.0721\topelement{0.0059} & 0.1537\topelement{0.0177} & 0.1125\topelement{0.0431} & \textbf{0.0318\topelement{0.0121}} \\
        & FairSHAP & & 0.6609\topelement{0.0106} & \textbf{0.0629\topelement{0.0091}} & \textbf{0.1326\topelement{0.0407}} & \textbf{0.0985\topelement{0.0603}} & 0.0452\topelement{0.0383} \\
        \midrule
        \multirow{4}{*}{Adult} 
        & Baseline & & 0.8722\topelement{0.0033} & 0.0315\topelement{0.0037} & 0.1805\topelement{0.0066} & 0.0735\topelement{0.0275} & \textbf{0.0275\topelement{0.0321}} \\
        & Ablation study 1 & & 0.8717\topelement{0.0031} & 0.0332\topelement{0.0020} & 0.1842\topelement{0.0048} & 0.0932\topelement{0.0186} & 0.0344\topelement{0.0228} \\
        & Ablation study 2 & & \textbf{0.8722\topelement{0.0016}} & 0.0278\topelement{0.0018} & 0.1816\topelement{0.0084} & 0.0761\topelement{0.0325} & 0.0442\topelement{0.0266} \\
        & FairSHAP & & 0.8692\topelement{0.0046} & \textbf{0.0273\topelement{0.0047}} & \textbf{0.1558\topelement{0.0130}} & \textbf{0.0393\topelement{0.0254}} & 0.0474\topelement{0.0319} \\
        \midrule
        \multirow{4}{*}{CensusIncome} 
        & Baseline & & 0.9377\topelement{0.0026} & 0.0767\topelement{0.0011} & 0.0012\topelement{0.0008} & 0.0012\topelement{0.0008} & 0.0796\topelement{0.0054} \\
        & Ablation study 1 & & 0.9377\topelement{0.0027} & 0.0765\topelement{0.0019} & 0.0010\topelement{0.0006} & 0.0011\topelement{0.0007} & 0.0797\topelement{0.0055} \\
        & Ablation study 2 & & 0.9378\topelement{0.0024} & 0.0744\topelement{0.0018} & 0.0010\topelement{0.0003} & 0.0010\topelement{0.0003} & 0.0796\topelement{0.0056} \\
        & FairSHAP & & \textbf{0.9381\topelement{0.0029}} & \textbf{0.0732\topelement{0.0020}} & \textbf{0.0008\topelement{0.0007}} & \textbf{0.0006\topelement{0.0008}} & \textbf{0.0794\topelement{0.0060}} \\
        \midrule
        \multirow{4}{*}{DefaultCredit} 
        & Baseline & & 0.8141\topelement{0.0059} & 0.0226\topelement{0.0019} & 0.0335\topelement{0.0072} & 0.0348\topelement{0.0211} & \textbf{0.0269\topelement{0.0127}} \\
        & Ablation study 1 & & 0.8149\topelement{0.0061} & 0.0221\topelement{0.0027} & 0.0327\topelement{0.0067} & 0.0431\topelement{0.0111} & 0.0369\topelement{0.0166} \\
        & Ablation study 2 & & \textbf{0.8150\topelement{0.0051}} & 0.0217\topelement{0.0031} & 0.0331\topelement{0.0070} & 0.0259\topelement{0.0135} & 0.0325\topelement{0.0137} \\
        & FairSHAP & & 0.8145\topelement{0.0050} & \textbf{0.0214\topelement{0.0026}} & \textbf{0.0306\topelement{0.0054}} & \textbf{0.0216\topelement{0.0164}} & 0.0289\topelement{0.0075} \\
        \bottomrule
    \end{tabular}
    }
\end{table*}
We set the threshold $T = 0.05$, meaning that we modify the feature values corresponding to Shapley values greater than 0.05, and the number of modifications under this threshold varies across datasets (see Appendix~\ref{app:threshold} for a detailed justification of this choice). As the number of modifications increases, we observe a consistent decrease in \gls{dr} across all datasets. Under the threshold of $T = 0.05$, FairSHAP is able to reduce \gls{dr} by up to 80\% for the German Credit dataset and by nearly 40\% for the COMPAS dataset. For the other two datasets, FairSHAP reduces the \gls{dr} by around 20\%. However, for the Census Income dataset, which contains a large number of features, the \gls{dr} reduction is only about 5\%.
This reduction in \gls{dr} is also accompanied by corresponding decreases in \gls{dp} and \gls{eo}, indicating that FairSHAP effectively enhances overall model fairness.\looseness=-1 

We also evaluated the performance of FairSHAP-modified data on other group fairness measures; detailed results can be found in Appendix~\ref{app1:FairSHAP}. 
Moreover, in datasets such as COMPAS, Census Income KDD, and Default Credit, the accuracy improves after applying modifications, demonstrating that FairSHAP can simultaneously enhance both fairness and predictive performance. In the remaining datasets, the accuracy experiences only a marginal decrease, indicating a balanced trade-off between fairness and model efficacy.

Furthermore, we conducted comparative experiments with other benchmark methods on the same datasets. The results, presented in Table~\ref{tab:fairness_comparison_extended}, show that FairSHAP improves all three fairness metrics (\gls{dr}, \gls{dp}, \gls{eo}) with minimal modifications. In some datasets, it reduces all fairness measures simultaneously. These improvements generally come with only a slight decrease in model accuracy, and in three of the datasets, they even lead to an accuracy increase. This comparative analysis confirms the reliability and effectiveness of FairSHAP in enhancing model fairness.  

Although {CorrelationRemover} performs well in terms of \gls{dr}, its mechanism relies on projection to identify the influence of the sensitive attribute on the non-sensitive attribute, subsequently adjusting the non-sensitive attribute to achieve a low DR value. However, this approach has several drawbacks:  
\textit{(1)} It modifies a large proportion of the data, exceeding 95\%, which significantly compromises data integrity.  
\textit{(2)} Its performance on \gls{dp} and \gls{eo} metrics is substantially worse than that of FairSHAP.  
\textit{(3)} It requires applying the same projection matrix to modify the test set.  

What's more, while {DisparateImpactRemover} can reduce certain values in terms of \gls{dp} and \gls{eo}, \textit{(1)} the extent of improvement across almost all metrics is generally inferior to that achieved by FairSHAP. \textit{(2)} It requires more modifications to the dataset compared to FairSHAP. \textit{(3)} It also necessitates applying the same transformations to the test set, which is not required by FairSHAP.

\subsection{Ablation study}
To evaluate the importance of each component in our method, we conducted below ablation studies:

\begin{itemize}[leftmargin=2em,topsep=4pt, itemsep=2pt,parsep=2pt]
    \item \textbf{Ablation study 1:} Instead of using the aforementioned $\mathcal{G}$ and $\mathcal{\tilde{G}}$, randomly sample a single data point from dataset $\mathcal{D}$ and replace its value in a specific column with a randomly selected value from the same column in other rows. The number of modifications performed should be approximately equal to that of FairSHAP.
    \item \textbf{Ablation study 2:} Using $\mathcal{G}$ and $\mathcal{\tilde{G}}$, match them via nearest neighbor, but instead of employing Shapley values to compute feature importance, directly replace values in $\mathcal{G}$ by randomly sampling from its reference data and replace values in $\mathcal{\tilde{G}}$ by randomly sampling from its corresponding reference data. The number of modifications performed should be approximately equal to that of FairSHAP.
\end{itemize}

The results, summarized in Table~\ref{tab:fairness_comparison_ablation}, show that the metrics in Ablation study 1 fluctuate slightly around the baseline, with overall performance slightly worse than the baseline. This deviation is due to the random modification of a small number of data points. In contrast, the metrics in Ablation study 2 show a slight improvement over the baseline, confirming that using matching and modifying values based on background data has a positive effect. This also highlights the importance of matching in FairSHAP. Finally, FairSHAP achieves overall improvements in \gls{dr}, \gls{dp}, and \gls{eo}, with some datasets even showing increased accuracy. These two ablation studies demonstrate that FairSHAP, based on matching and Shapley values, is effective for fairness-aware data augmentation.  

However, when dealing with large datasets, the gap between FairSHAP and the Ablation study becomes smaller. We believe this is because one-hot encoding, used during preprocessing, generates a large number of features. As SHAP's kernel method relies on a random baseline to reduce computational complexity, identifying important features and making adjustments becomes more challenging.

\section{Conclusion}
In this paper, we introduced FairSHAP, an innovative preprocessing framework that leverages Shapley value attribution to enhance fairness in ML models. FairSHAP effectively identifies and systematically adjusts fairness-critical instances using interpretable feature importance measures and instance-level matching across sensitive groups. Our approach significantly reduces fairness metrics such as \gls{dr}, \gls{dp} and \gls{eo}  across various datasets. Crucially, FairSHAP achieves these improvements while perturbing the data as minimally as possible, largely preserving data fidelity and, in certain cases, even enhancing predictive accuracy. As a transparent and model-agnostic method, FairSHAP integrates into existing machine learning pipelines, providing actionable insights into bias sources and fostering accountable and equitable AI systems.
\paragraph{Future work} In our future work, we target: 
\textit{(1)} Finding a new matching method relationship in addition to nearest-neighbor matching;
\textit{(2)} Accelerating FairSHAP by integrating more efficient Shapley value estimation methods, in order to overcome computational complexity limitations (see Appendix~\ref{app:complexity} and Appendix~\ref{app:appendix_shap_experiments}).

\bibliographystyle{plainnat}
\bibliography{references}

\clearpage
\appendix
\onecolumn
\appendixpage 

{
\small
\setcounter{tocdepth}{1}
\startcontents[sections]
\printcontents[sections]{l}{1}{}
}

\section{Fairness metrics across datasets}
\label{app1:FairSHAP}
\begin{figure}[h!]
\vspace{-3.5mm}
    \centering
    \includegraphics[width=.9921\linewidth]{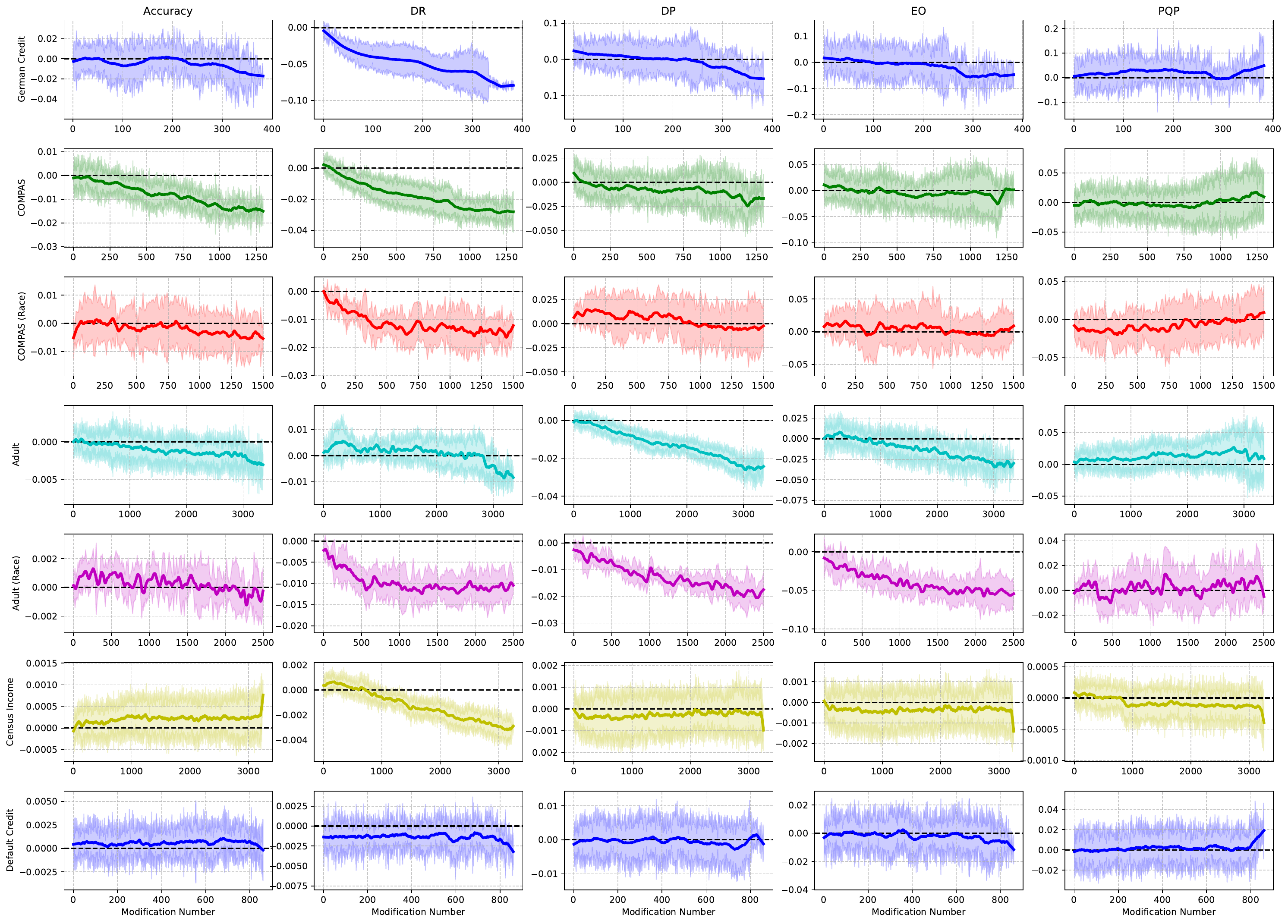}
    \vspace{-1.45em}
    \caption{Improvement in fairness metrics (Accuracy, DR, DP, EO, and PQP) as the number of modifications increases using FairSHAP on various datasets. Solid lines represent the mean, and shaded areas indicate the standard deviation from 5-fold cross-validation. The $x$-axis shows the number of modifications taken, up to the maximum required under a given threshold $T$, where $T=0.05$.}
    \label{fig:all datasets use FairSHAP}
\end{figure}

\section{Definition of group fairness measures}
\label{definition_group_fairness}

\paragraph{Notation and setup.}
Let $\mathcal{X}$ denote the feature space, and each instance be written as $\mathbf{x}=(\check{\mathbf{x}}, A)$, where $\check{\mathbf{x}} \in \mathcal{X}_{\mathcal{F}\setminus\{A\}}$ are the non-sensitive features and $A\in\{0,1\}$ is the sensitive attribute (0 and 1 representing the unprivileged and privileged groups, respectively).  
We denote by $\mathbb{P}_{\mathcal{D}}(\cdot)$ the probability with respect to the underlying data distribution $\mathcal{D}$ over $(\mathcal{X},\mathcal{Y})$.  

\paragraph{Demographic parity (DP).}
DP requires that the probability of a positive prediction be independent of the sensitive attribute:
\begin{equation}
\label{eq:dp}
L_{\text{DP}}(f) = 
\big| \mathbb{P}_{\mathcal{D}}\big(f(\mathbf{x})=1 \mid A=0\big) 
- \mathbb{P}_{\mathcal{D}}\big(f(\mathbf{x})=1 \mid A=1\big)\big| 
\leq \varepsilon \,.
\end{equation}
A model violating this condition exhibits \emph{disparate impact}, since the chance of receiving a favorable outcome differs between groups.

\paragraph{Equality of opportunity (EO).}
EO requires that, among truly qualified individuals ($y=1$), the chance of a positive prediction does not depend on the sensitive attribute:
\begin{equation}
L_{\text{EO}}(f) =
\big| \mathbb{P}_{\mathcal{D}}\big(f(\mathbf{x})=1 \mid A=0, y=1\big) 
- \mathbb{P}_{\mathcal{D}}\big(f(\mathbf{x})=1 \mid A=1, y=1\big)\big|
\leq \varepsilon \,.
\end{equation}
Failure to satisfy EO indicates \emph{disparate treatment}, disadvantaging some groups despite equal qualifications.

\paragraph{Predictive quality parity (PQP).}
PQP requires that the reliability of positive predictions be consistent across groups:
\begin{equation}
L_{\text{PQP}}(f) =
\big| \mathbb{P}_{\mathcal{D}}\big(y=1 \mid f(\mathbf{x})=1, A=0\big) 
- \mathbb{P}_{\mathcal{D}}\big(y=1 \mid f(\mathbf{x})=1, A=1\big)\big|
\leq \varepsilon \,.
\end{equation}
Violation of PQP implies that the correctness of positive predictions differs across groups, which is undesirable in high-stakes domains such as healthcare or finance.

\section{DR Reduction Achieved by FairSHAP}
We now establish the theoretical foundation of FairSHAP by analyzing its impact on the individual fairness measure, \gls{dr}.  
Our analysis begins by showing that $L_{\text{DR}}$ admits a natural decomposition into feature-wise Shapley contributions, which allows us to identify fairness-critical features.  
Building on this, we derive a provable lower bound on the reduction in $L_{\text{DR}}$ that can be achieved by selectively modifying such features.

\paragraph{Notation and setup.}
Let $\mathcal{X}$ denote the feature space, $\mathcal{F}$ denotes full set of features, and each instance be written as $\mathbf{x}=(\check{\mathbf{x}}, A)$, where $\check{\mathbf{x}} \in \mathcal{X}_{\mathcal{F}\setminus\{A\}}$ are the non-sensitive features and $A\in\{0,1\}$ is the sensitive attribute (0 and 1 representing the unprivileged and privileged groups, respectively).  
We denote by $\mathbb{P}_{\mathcal{D}}(\cdot)$ the probability with respect to the underlying data distribution $\mathcal{D}$ over $(\mathcal{X},\mathcal{Y})$.  

\begin{lemma}[Shapley decomposition of individual DR]\label{lem:shapley-decomp}
Let $\mathbf{x} = (\check{\mathbf{x}}, A)$ be an instance, where $\check{\mathbf{x}} \in \mathcal{X}_{\mathcal{F}\setminus\{A\}}$ 
denotes the non-sensitive features and $A \in \{0,1\}$ is the sensitive attribute.  

The individual discriminative risk is defined as
\[
L_{\myDR}(f, \mathbf{x}) 
= \big| f(\check{\mathbf{x}},A=0) - f(\check{\mathbf{x}},A=1) \big| \,.
\]

Fix a reference instance $\mathbf{r}$ drawn from the reference set $\mathcal{B}$ and consider the coalition game
\[
v^{({\mathbf{x}})}_{\myDR}(S) 
= L_{\myDR}\big({\mathbf{x}}_S;\, {\mathbf{r}}_{\mathcal{F}\setminus S}\big) \,,
\qquad S \subseteq [d] \,.
\]
Let $\phi_k({\mathbf{x}})$ denote the Shapley value of feature $k$ in this game and define the baseline term as
\[
\phi_0(\mathbf{x}) := v^{(\mathbf{x})}_{\myDR}(\emptyset) \,.
\]
Then, for the grand coalition $S=[d]$, we have
\[ 
L_{\myDR}(f,{\mathbf{x}}) = \phi_0({\mathbf{x}}) + \sum_{k=1}^d \phi_k({\mathbf{x}}) \,.
\]
\end{lemma}

\begin{proof}
We recall that the Shapley value $\phi_j$ for feature $j$ satisfies the \emph{efficiency} axiom, i.e.,
\[ \textstyle
\sum_{j=1}^d \phi_j(\mathbf{x}) = v^{(\mathbf{x})}_{\myDR}([d]) - v^{(\mathbf{x})}_{\myDR}(\emptyset) \,,
\]
which states that the sum of the contributions of all features equals the total value of the grand coalition relative to the empty coalition.  

In our setting, for the grand coalition $S=[d]$, we have
\[
v^{(\mathbf{x})}_{\myDR}([d]) = L_{\myDR}(f,\mathbf{x}) \,,
\]
and for the empty coalition $S=\emptyset$, all features are taken from the reference instance, hence
\[
v^{(\mathbf{x})}_{\myDR}(\emptyset) = L_{\myDR}(f,\mathbf{r}) \,.
\]
Rearranging the efficiency identity and using the definition $\phi_0(\mathbf{x}) := v^{(\mathbf{x})}_{\myDR}(\emptyset)$ yields
\[
L_{\myDR}(f,\mathbf{x})
= v^{(\mathbf{x})}_{\myDR}([d])
= v^{(\mathbf{x})}_{\myDR}(\emptyset) + \sum_{j=1}^d \phi_j(\mathbf{x})
= \phi_0(\mathbf{x}) + \sum_{j=1}^d \phi_j(\mathbf{x}) \,.
\]
\end{proof}

Lemma~\ref{lem:shapley-decomp} provides a rigorous way to decompose $L_{\text{DR}}$ into additive contributions of individual features.  
Intuitively, features with large Shapley values exert disproportionate influence on model predictions, and are thus most likely to drive individual-level unfairness.  
We therefore define these as {fairness-critical features. By neutralizing their effect through FairSHAP, we can guarantee a measurable reduction in $L_{\myDR}$.

\begin{theorem}[Instance-wise DR reduction achieved by FairSHAP]\label{thm:dr-reduction}
Fix a threshold $T > 0$ and define the set of fairness-critical features as
\[
S_T({\mathbf{x}}) = \{\, k \in [d] \;\mid\; \phi_k({\mathbf{x}}) \ge T \,\} \,.
\]
Let ${\mathbf{x}}_{\mathrm{after}}$ denote the feature vector obtained from ${\mathbf{x}}$ after FairSHAP replaces every feature $k \in S_T({\mathbf{x}})$ by its value in the matched reference ${\mathbf{r}}$.  
Define the interaction residual term as
\[
\mathcal{I}({\mathbf{x}}) \;:=\; 
L_{\myDR}(f,{\mathbf{x}}_{\mathrm{after}})
-\Bigg(L_{\myDR}(f,{\mathbf{x}}_{\mathrm{before}})-\sum_{k\in S_T({\mathbf{x}})} \phi_k({\mathbf{x}})\Bigg).
\]
Then
\[ 
L_{\myDR}(f,{\mathbf{x}}_{\mathrm{after}})
=
L_{\myDR}(f,{\mathbf{x}}_{\mathrm{before}})
-
\sum_{k\in S_T({\mathbf{x}})} \phi_k({\mathbf{x}})
+ \mathcal{I}({\mathbf{x}})
\;\;\le\;\;
L_{\myDR}(f,{\mathbf{x}}_{\mathrm{before}})
-
T\, |S_T({\mathbf{x}})|
+ \mathcal{I}({\mathbf{x}}) \,,
\]
where \(\mathcal{I}({\mathbf{x}})\) denotes the interaction residual term.
In particular, if $\mathcal{I}(\mathbf{x}) \le T|S_T(\mathbf{x})|$, then $L_{\myDR}(f,\mathbf{x}_{\mathrm{after}})\le L_{\myDR}(f,\mathbf{x}_{\mathrm{before}})$, i.e., the DR does not increase.
\end{theorem}

\begin{proof}
By Lemma~\ref{lem:shapley-decomp},
\[ 
L_{\myDR}(f,{\mathbf{x}}_{\mathrm{before}})
= \phi_0({\mathbf{x}}) 
+ \sum_{k \in S_T({\mathbf{x}})} \phi_k({\mathbf{x}})
+ \sum_{k \notin S_T({\mathbf{x}})} \phi_k({\mathbf{x}}) \,.
\]
By the definition of $\mathcal{I}(\mathbf{x})$ above (which accounts for non-additive interactions when replacing the subset $S_T(\mathbf{x})$),
\[ 
L_{\myDR}(f,{\mathbf{x}}_{\mathrm{after}})
= L_{\myDR}(f,{\mathbf{x}}_{\mathrm{before}})
- \sum_{k \in S_T({\mathbf{x}})} \phi_k({\mathbf{x}}) + \mathcal{I}(\mathbf{x}) \,.
\]
Since $\phi_k(x) \ge T$ for all $k \in S_T(x)$, by the definition of $S_T(x)$ we have
\[ 
\sum_{k \in S_T(x)} \phi_k(x) \;\ge\; T\,|S_T(x)| \,.
\]
Substituting completes the proof.
\end{proof}

Theorem~\ref{thm:dr-reduction} provides a per-instance bound on the post-modification DR \emph{up to the residual} $\mathcal{I}(\mathbf{x})$. Therefore, a strict improvement guarantee requires controlling $\mathcal{I}(\mathbf{x})$, e.g., via a separate bound in Proposition~\ref{prop:residual_bound}.
More precisely, the theorem implies
\[
L_{\myDR}(f,\mathbf{x}_{\mathrm{after}})\;\le\;L_{\myDR}(f,\mathbf{x}_{\mathrm{before}})\;-\;T|S_T(\mathbf{x})|\;+\;\mathcal{I}(\mathbf{x}) \,,
\]
so a \emph{non-increase} (and hence an improvement) is guaranteed whenever $\mathcal{I}(\mathbf{x}) \le T|S_T(\mathbf{x})|$, in which case the reduction is at least $T|S_T(\mathbf{x})|-\mathcal{I}(\mathbf{x})$.
Accordingly, we refrain from claiming a uniform per-feature reduction of at least $T$ without an explicit bound on $\mathcal{I}(\mathbf{x})$.
This establishes a principled lower bound on fairness improvement (assuming \(|\mathcal{I}(\mathbf{x})|\) is small via local matching): the more fairness-critical features are neutralized, the greater the guaranteed reduction in individual-level unfairness.

\vspace{0.5em} 

\begin{proposition}[Bound on Interaction Residual]\label{prop:residual_bound}
Let $L(\mathbf{x}) := L_{\myDR}(f,\mathbf{x})$ denote the instance-wise DR objective used in Theorem~\ref{thm:dr-reduction}. 
Since the absolute-value DR may be non-differentiable when $f(\check{\mathbf{x}},0)=f(\check{\mathbf{x}},1)$, we assume either (i) a smooth surrogate $L_\delta$ is used (e.g., $L_\delta(\mathbf{x})=\sqrt{(f(\check{\mathbf{x}},0)-f(\check{\mathbf{x}},1))^2+\delta}$), or (ii) we restrict to a neighborhood where the sign of $f(\check{\mathbf{x}},0)-f(\check{\mathbf{x}},1)$ is constant, so that $L$ is twice differentiable locally.

Since FairSHAP uses NN or Optimal Transport to select a reference \(\mathbf{r}\) from the local neighborhood of $\mathbf{x}_{\mathrm{before}}$, let $\epsilon := \|\mathbf{x}_{\mathrm{before}} - \mathbf{r}\|$ be the matching distance. 
Assume $L$ is twice differentiable on the closed ball $B(\mathbf{x}_{\mathrm{before}},\epsilon):=\{\mathbf{z}:\|\mathbf{z}-\mathbf{x}_{\mathrm{before}}\|\le \epsilon\}$ and its Hessian is bounded there:
\[
\sup_{\mathbf{z}\in B(\mathbf{x}_{\mathrm{before}},\epsilon)}\big\|H_L(\mathbf{z})\big\|_{\mathrm{op}} \le M \,,
\]
so that $\nabla L$ is $M$-Lipschitz on $B(\mathbf{x}_{\mathrm{before}},\epsilon)$.

If $\mathbf{x}_{\mathrm{after}}$ is obtained by replacing the subset $S_T(\mathbf{x}_{\mathrm{before}})$ with reference coordinates, define $\Delta:=\mathbf{x}_{\mathrm{after}}-\mathbf{x}_{\mathrm{before}}$. Then $\|\Delta\|\le \epsilon$.
Then, the interaction residual is bounded by a second-order term:
\[
|\mathcal{I}(\mathbf{x}_{\mathrm{before}})| \le \bigg(\frac{M}{2} + M\sqrt{|S_T(\mathbf{x}_{\mathrm{before}})|}\bigg) \cdot \epsilon^2 \,,
\]
hence in particular $|\mathcal{I}(\mathbf{x}_{\mathrm{before}})|\le C\epsilon^2$ for $C:=\frac{M}{2}+M\sqrt{d}$ when the dimension $d$ is fixed.
Consequently, as $\epsilon \to 0$, the residual $\mathcal{I}(\mathbf{x}_{\mathrm{before}})$ is second-order. Moreover, if $\|\nabla L(\mathbf{x}_{\mathrm{before}})\|_\infty$ is locally bounded, then each first-order term scales as $O(\epsilon)$, so the residual vanishes faster than the leading contributions, justifying the approximation in Theorem~\ref{thm:dr-reduction}.
\end{proposition}

\begin{proof}
Let $\Delta:=\mathbf{x}_{\mathrm{after}}-\mathbf{x}_{\mathrm{before}}$. By the multivariate second-order Taylor theorem with Lagrange remainder, there exists $t^\star\in(0,1)$ such that
\[
L(\mathbf{x}_{\mathrm{after}})
= L(\mathbf{x}_{\mathrm{before}}) + \nabla L(\mathbf{x}_{\mathrm{before}})^\top \Delta
+ \frac12\,\Delta^\top H_L(\mathbf{x}_{\mathrm{before}}+t^\star\Delta)\,\Delta \,.
\]
Rearranging gives
\[
L(\mathbf{x}_{\mathrm{before}})-L(\mathbf{x}_{\mathrm{after}})
= -\nabla L(\mathbf{x}_{\mathrm{before}})^\top \Delta
-\frac12\,\Delta^\top H_L(\mathbf{x}_{\mathrm{before}}+t^\star\Delta)\,\Delta \,.
\]
Taking absolute values and using $\|H_L(\cdot)\|_{\mathrm{op}}\le M$ on $B(\mathbf{x}_{\mathrm{before}},\epsilon)$ yields
\[
\Big|\,\big(L(\mathbf{x}_{\mathrm{before}})-L(\mathbf{x}_{\mathrm{after}})\big)
- \big(-\nabla L(\mathbf{x}_{\mathrm{before}})^\top \Delta\big)\,\Big|
\le \frac{M}{2}\,\|\Delta\|^2 \,.
\]

Next we relate the linear term to the Shapley sum under the baseline-replacement game
$v^{(\mathbf{x}_{\mathrm{before}})}(U)=L\big((\mathbf{x}_{\mathrm{before}})_U;\,\mathbf{r}_{[d]\setminus U}\big)$.
Define the hybrid point $\mathbf{z}_U$ by $(\mathbf{z}_U)_i=(\mathbf{x}_{\mathrm{before}})_i$ if $i\in U$ and $(\mathbf{z}_U)_i=r_i$ otherwise. Then
\[
\|\mathbf{z}_U-\mathbf{x}_{\mathrm{before}}\|^2=\sum_{i\notin U}(r_i-(\mathbf{x}_{\mathrm{before}})_i)^2\le \sum_{i=1}^d(r_i-(\mathbf{x}_{\mathrm{before}})_i)^2=\epsilon^2 \,,
\]
so $\mathbf{z}_U\in B(\mathbf{x}_{\mathrm{before}},\epsilon)$ for all $U\subseteq[d]$. Since $B(\mathbf{x}_{\mathrm{before}},\epsilon)$ is convex, the segment between any two hybrid points (hence the mean-value point $\boldsymbol{\xi}_{S,k}$ below) also lies in $B(\mathbf{x}_{\mathrm{before}},\epsilon)$.

For any $k\in S_T(\mathbf{x}_{\mathrm{before}})$, the Shapley value is a weighted average of marginal increments
$m_{S,k}:=v^{(\mathbf{x}_{\mathrm{before}})}(S\cup\{k\})-v^{(\mathbf{x}_{\mathrm{before}})}(S)$ over $S\subseteq[d]\setminus\{k\}$.
Fix such an $S$ and define $g(t):=L(\mathbf{z}_S+t((\mathbf{x}_{\mathrm{before}})_k-r_k)\mathbf{e}_k)$. By the mean value theorem, 
$m_{S,k}=g(1)-g(0)=\partial_k L(\boldsymbol{\xi}_{S,k})\,((\mathbf{x}_{\mathrm{before}})_k-r_k)$ for some $\boldsymbol{\xi}_{S,k}$ on the segment between the two hybrids.
Since $\boldsymbol{\xi}_{S,k}\in B(\mathbf{x}_{\mathrm{before}},\epsilon)$ and $\nabla L$ is $M$-Lipschitz on this ball, we have
\[
|\partial_k L(\boldsymbol{\xi}_{S,k})-\partial_k L(\mathbf{x}_{\mathrm{before}})|
\le \|\nabla L(\boldsymbol{\xi}_{S,k})-\nabla L(\mathbf{x}_{\mathrm{before}})\|
\le M\|\boldsymbol{\xi}_{S,k}-\mathbf{x}_{\mathrm{before}}\|
\le M\epsilon \,.
\]
Therefore,
\[
|m_{S,k}-\partial_k L(\mathbf{x}_{\mathrm{before}})\big((\mathbf{x}_{\mathrm{before}})_k-r_k\big)|
\le M\epsilon \big|(\mathbf{x}_{\mathrm{before}})_k-r_k\big| \,.
\]
Averaging over $S$ (i.e., over the Shapley weights) yields
\[
|\phi_k(\mathbf{x}_{\mathrm{before}})-\partial_k L(\mathbf{x}_{\mathrm{before}})\big((\mathbf{x}_{\mathrm{before}})_k-r_k\big)|
\le M\epsilon \big|(\mathbf{x}_{\mathrm{before}})_k-r_k\big| \,.
\]

Summing over $k\in S_T(\mathbf{x}_{\mathrm{before}})$ and using Cauchy--Schwarz gives
\[
\begin{aligned}
\Bigg|\sum_{k\in S_T(\mathbf{x}_{\mathrm{before}})}\phi_k(\mathbf{x}_{\mathrm{before}})
-\sum_{k\in S_T(\mathbf{x}_{\mathrm{before}})}\partial_k L(\mathbf{x}_{\mathrm{before}})\big((\mathbf{x}_{\mathrm{before}})_k-r_k\big)\Bigg| 
&\le M\epsilon \sum_{k\in S_T(\mathbf{x}_{\mathrm{before}})}\big|(\mathbf{x}_{\mathrm{before}})_k-r_k\big| \\
&\le M\epsilon\sqrt{|S_T(\mathbf{x}_{\mathrm{before}})|}\,\|\Delta\| \\
&\le M\sqrt{|S_T(\mathbf{x}_{\mathrm{before}})|}\,\epsilon^2 \,.
\end{aligned}
\]

Finally, note that $-\nabla L(\mathbf{x}_{\mathrm{before}})^\top\Delta
=\sum_{k\in S_T(\mathbf{x}_{\mathrm{before}})}\partial_k L(\mathbf{x}_{\mathrm{before}})\big((\mathbf{x}_{\mathrm{before}})_k-r_k\big)$, and recall the residual definition from Theorem~\ref{thm:dr-reduction}:
\[
\mathcal{I}(\mathbf{x}_{\mathrm{before}})
:= L(\mathbf{x}_{\mathrm{after}})
-\Bigg(L(\mathbf{x}_{\mathrm{before}})-\sum_{k\in S_T(\mathbf{x}_{\mathrm{before}})}\phi_k(\mathbf{x}_{\mathrm{before}})\Bigg) \,.
\]
Combining the two bounds yields
\[
|\mathcal{I}(\mathbf{x}_{\mathrm{before}})|
\le \frac{M}{2}\|\Delta\|^2 + M\sqrt{|S_T(\mathbf{x}_{\mathrm{before}})|}\,\epsilon^2
\le \bigg(\frac{M}{2}+M\sqrt{|S_T(\mathbf{x}_{\mathrm{before}})|}\bigg)\epsilon^2 \,,
\]
and using $\|\Delta\|\le \epsilon$ completes the proof.
\end{proof}

\section{Bounding demographic parity by discriminative risk}\label{app3:dr-dp}

Beyond individual fairness, we show that reducing individuallevel DR also controls group
level statistical disparities such as DP. Intuitively, features that are fairnesscritical at the
individual level tend to correlate with (or encode) the sensitive attribute, so neutralizing
their effect reduces systematic differences between demographic groups. This section
formalizes that link.

\paragraph{Notation and setup.}
Let $\mathcal{X}$ denote the feature space, $\mathcal{F}$ denotes full set of features, and each instance be written as $\mathbf{x}=(\check{\mathbf{x}}, A)$, where $\check{\mathbf{x}} \in \mathcal{X}_{\mathcal{F}\setminus\{A\}}$ are the non-sensitive features and $A\in\{0,1\}$ is the sensitive attribute (0 and 1 representing the unprivileged and privileged groups, respectively). For a classifier
$f:\mathcal{X}_{\mathcal{F} \setminus \{A\}} \times\{0,1\}\!\to\![0,1]$, we write $f(\check{\mathbf{x}},A=a)=1$, if the instance is predicted positive.

\paragraph{Discriminative risk (DR)}
For a single instance $\mathbf{x} = (\check{\mathbf{x}},A)$,
\[
  L_\myDR(f,\mathbf{x})
  = \big|\,f(\check{\mathbf{x}},A=0)-f(\check{\mathbf{x}},A=1)\,\big| 
  \,.
\]
For any distribution $D$ over $\check{\mathbf{x}}$ (e.g., the empirical mixture of the two groups), the population DR is
\begin{equation}
\label{distribution_Dr}
  L_\myDR(f;D)
  = \mathbb{E}_{\check{\mathbf{x}}\sim D}\big[L_\myDR(f,\mathbf{x})\big]
  = \mathbb{E}_{\check{\mathbf{x}}\sim D}\big|\,f(\check{\mathbf{x}},A=0)-f(\check{\mathbf{x}},A=1)\,\big|
  \,.
\end{equation}

\paragraph{Demographic parity (DP)}
As introduced earlier in equation~\eqref{eq:dp}, the classical definition of demographic parity is based on the difference in positive prediction rates between the two sensitive groups. 
For clarity, we rewrite it here and denote it with a superscript ``decision'':
\begin{equation}\label{eq:dp-classical}
L_{\text{DP}}^{\text{decision}}(f) 
= \big| \mathbb{P}_{\mathcal{D}}\big(f(\mathbf{x})=1 \mid A=0\big)
- \mathbb{P}_{\mathcal{D}}\big(f(\mathbf{x})=1 \mid A=1\big)\big|
\,.
\end{equation}

In practice, however, modern models typically output continuous scores 
(e.g., probabilities in $[0,1]$), and the decision-based formulation 
depends on the choice of a threshold to binarize predictions. 
This threshold dependence makes the measure unstable and less suitable for 
optimization. To address this, we adopt a score-based relaxation of demographic parity, 
which compares the average prediction scores between the two groups. 
Let $D_0$ denote the conditional distribution of $\check{\mathbf{x}}$ given $A=0$, 
and $D_1$ denote the conditional distribution of $\check{\mathbf{x}}$ given $A=1$. 
The score-based DP gap is defined as
\begin{equation}\label{eq:dp-score}
L_{\text{DP}}^{\text{score}}(f)
= \big|\,\mathbb{E}_{\check{\mathbf{x}}\sim D_0}[f(\check{\mathbf{x}},A=0)]
- \mathbb{E}_{\check{\mathbf{x}}\sim D_1}[f(\check{\mathbf{x}},A=1)]\,\big|
\,.
\end{equation}

\subsection{Identical conditional distributions}
\begin{theorem}[Identical-distribution case]
\label{thm:dp-dr-identical}
If $D_0=D_1=D$, then for any classifier $f$,
\[
  L_{\mathrm{DP}}^{\text{score}}(f) \;\le\; L_\myDR(f;D)
  \,.
\]
\end{theorem}

\begin{proof}
Under $D_0=D_1=D$,
%
\[ \begin{split}
L_{\mathrm{DP}}^{\text{score}}(f)
  &= \big|\,\mathbb{E}_{\check{\mathbf{x}}\sim D}[f(\check{\mathbf{x}},A=0)]
- \mathbb{E}_{\check{\mathbf{x}}\sim D}[f(\check{\mathbf{x}},A=1)]\,\big| \\
  &= \big|\,\mathbb{E}_{\check{\mathbf{x}}\sim D}\big[f(\check{\mathbf{x}},A=0)-f(\check{\mathbf{x}},A=1)\big]\,\big| \\
  &\leq\; 
  \mathbb{E}_{\check{\mathbf{x}}\sim D}\big|f(\check{\mathbf{x}},A=0)-f(\check{\mathbf{x}},A=1)\big| \\
  &= L_\myDR(f;D) \,,
\end{split} \]
where we used Jensen’s inequality.
\end{proof}

\subsection{General case: distribution shift between groups}
We recall that $D_0$ denotes the conditional distribution of $\check{\mathbf{x}}$ given $A=0$, 
and $D_1$ denotes the conditional distribution of $\check{\mathbf{x}}$ given $A=1$.  
We assume throughout this section that
\[
    f(\check{\mathbf{x}},A)\in[0,1] \,,
    \quad\text{for }A\in\{0,1\} \,.
\]

\begin{definition}[Total variation distance]
Let $D_0,D_1$ be two probability distributions over the non-sensitive feature space 
$\check{\mathbf{x}} \in \mathcal{X}_{\mathcal{F}\setminus\{A\}}$. 
The \emph{total-variation (TV) distance} between them is defined as
\begin{equation}\label{eq:def-tv}
\mathrm{TV}(D_0,D_1) 
\;=\; \sup_{B\subseteq \mathcal{X}_{\mathcal{F}\setminus\{A\}}} 
\;\big|\,D_0(B)-D_1(B)\,\big| \,,
\end{equation}
where the supremum $\sup$ is taken over all measurable subsets 
of the non-sensitive feature space. 
Equivalently, for any measurable $h:\mathcal{X}_{\mathcal{F}\setminus\{A\}}\to[0,1]$,
\begin{equation}\label{eq:tv-bound}
\big|\,\mathbb{E}_{\check{\mathbf{x}}\sim D_0}[h(\check{\mathbf{x}})]
-\mathbb{E}_{\check{\mathbf{x}}\sim D_1}[h(\check{\mathbf{x}})]\,\big|
\;\le\;\mathrm{TV}(D_0,D_1) \,.
\end{equation}
\end{definition}

\begin{theorem}[General case: distribution shift]
\label{thm:dp-dr-general}
For any classifier $f$ with outputs in $[0,1]$,
\begin{equation}
  L_{\mathrm{DP}}^{\text{score}}(f)
  \;\le\;
  \min\{\,L_\myDR(f;D_0),\,L_\myDR(f;D_1)\,\}
  \;+\; \mathrm{TV}(D_0,D_1) \,.
\end{equation}
\end{theorem}

\begin{proof}
Starting from the definition of the score-based DP gap:
\[
  L_{\mathrm{DP}}^{\text{score}}(f)
  = \big|\,\mathbb{E}_{\check{\mathbf{x}}\sim D_0}[f(\check{\mathbf{x}},0)]
  - \mathbb{E}_{\check{\mathbf{x}}\sim D_1}[f(\check{\mathbf{x}},1)]\,\big|
  \,.
\]

Introduce a coupling $\pi(\check{\mathbf{x}}_0,\check{\mathbf{x}}_1)$ 
with marginals $D_0$ and $D_1$. Then
\[
  L_{\mathrm{DP}}^{\text{score}}(f)
  = \big|\,\mathbb{E}_{\pi}\big[f(\check{\mathbf{x}}_0,0) - f(\check{\mathbf{x}}_1,1)\big]\,\big|
  \,.
\]

Add and subtract $f(\check{\mathbf{x}}_0,1)$ inside the expectation:
\[
  f(\check{\mathbf{x}}_0,0) - f(\check{\mathbf{x}}_1,1)
  =
  \big(f(\check{\mathbf{x}}_0,0) - f(\check{\mathbf{x}}_0,1)\big)
  + \big(f(\check{\mathbf{x}}_0,1) - f(\check{\mathbf{x}}_1,1)\big)
  \,.
\]

By the triangle inequality,
\[
  L_{\mathrm{DP}}^{\text{score}}(f)
  \;\le\;
  \mathbb{E}_{\pi}\big|f(\check{\mathbf{x}}_0,0) - f(\check{\mathbf{x}}_0,1)\big|
  + \big|\mathbb{E}_{\pi}[f(\check{\mathbf{x}}_0,1) - f(\check{\mathbf{x}}_1,1)]\big|
  \,.
\]

The first term gives $L_\myDR(f;D_0)$, and by symmetry the same argument 
with $f(\check{\mathbf{x}}_1,0)$ gives $L_\myDR(f;D_1)$. 
Thus we can bound by the smaller of the two:
\[
  \min\{\,L_\myDR(f;D_0),\,L_\myDR(f;D_1)\,\} \,.
\]

For the second term, define $h(\check{\mathbf{x}})=f(\check{\mathbf{x}},1)\in[0,1]$. Then
\[
  \mathbb{E}_{\pi}\big[f(\check{\mathbf{x}}_0,1) - f(\check{\mathbf{x}}_1,1)\big]
  = \mathbb{E}_{\check{\mathbf{x}}\sim D_0}[h(\check{\mathbf{x}})]
  - \mathbb{E}_{\check{\mathbf{x}}\sim D_1}[h(\check{\mathbf{x}})] \,,
\]
which is bounded by $\mathrm{TV}(D_0,D_1)$ via~\eqref{eq:tv-bound}. 

Therefore,
\[
  L_{\mathrm{DP}}^{\text{score}}(f) 
  \;\le\; \min\{\,L_\myDR(f;D_0),\,L_\myDR(f;D_1)\,\} + \mathrm{TV}(D_0,D_1) \,.
\]
\end{proof}

\paragraph{Implications.}
\begin{itemize}[leftmargin=2em,topsep=4pt, itemsep=2pt,parsep=2pt]
    \item If $D_0=D_1$, then $\mathrm{TV}(D_0,D_1)=0$, and the DP gap is bounded directly by the population discriminative risk, i.e., $L_\myDR(f;D)$; equivalently, since $D_0=D_1=D$, one can also write the bound in terms of either group-specific risk $L_\myDR(f;D_0)$ or $L_\myDR(f;D_1)$.

  \item If $D_0\neq D_1$, the theorem shows that reducing the discriminative risk of either group still tightens an upper bound on the DP gap. However, an irreducible term $\mathrm{TV}(D_0,D_1)$ remains due to inherent distribution shift rather than model bias.
\end{itemize}

\section{Empirical Validation of Matching Distance}
\label{app:matching_distance}

Proposition~\ref{prop:residual_bound} relies on the premise that the matching distance
\(\epsilon = \lVert \mathbf{x} - \mathbf{r} \rVert\) remains sufficiently small, so that the
higher-order interaction residual \(\mathcal{I}(\mathbf{x})\) is suppressed relative to the
first-order Shapley terms. In addition, since our method relies on distance-based matching,
we validate that the selected matched pairs are indeed closer than random pairings even in
high-dimensional and potentially sparse feature spaces, where Euclidean distances can be
less reliable.

For each query instance \(\mathbf{x}\), we select a reference \(\mathbf{r}\) using our
matching strategy (Nearest Neighbor matching with \(k=1\); we use the same feature space
and preprocessing as in the main experiments). We then compute the matching distance
\(\epsilon(\mathbf{x}) = \lVert \mathbf{x} - \mathbf{r} \rVert_2\). As a baseline, we also
pair each \(\mathbf{x}\) with a randomly sampled reference \(\tilde{\mathbf{r}}\) from the
reference pool and compute \(\lVert \mathbf{x} - \tilde{\mathbf{r}} \rVert_2\). We report
summary statistics over \(N=5{,}000\) query instances.

Table~\ref{tab:matching_quality} shows that matched pairs have substantially smaller
Euclidean distances than random pairs across two large and high-dimension datasets, Adult and Census Income KDD.
The differences are statistically significant (\(p<0.001\)) with large effect sizes (Cohen's
\(d=2.03\) on Adult and \(d=2.51\) on Census KDD), indicating that our matching strategy
consistently retrieves structurally similar references within the local neighborhood of
\(\mathbf{x}\). Combined with the quadratic scaling
\(\lvert \mathcal{I}(\mathbf{x}) \rvert \le C \epsilon^2\) in
Proposition~\ref{prop:residual_bound}, these results support the practical validity of the
approximation used in Theorem~\ref{thm:dr-reduction}.

\begin{table}[t]
    \centering\vspace{-2mm}
    \caption{\textbf{Quantitative validation of matching quality.} Comparison of Euclidean
    distances in feature space between matched pairs (via KNN, \(k=1\)) and random pairs
    across two datasets (\(N=5{,}000\)). All differences are statistically significant with
    \(p<0.001\). The \(t\)-statistic and Cohen's \(d\) compare matched vs.\ random distances.}
    \label{tab:matching_quality}
    \vspace{0.25em}
    \scalebox{.91}{
    \begin{tabular}{l l c c c c}
        \toprule
        \textbf{Dataset} & \textbf{Group} & \textbf{Mean Dist.} & \textbf{Std. Dev.} & \textbf{\(t\)-statistic} & \textbf{Cohen's \(d\)} \\
        \midrule
        \multirow{2}{*}{Adult} 
            & Matched & 1.56 & 0.75 & \multirow{2}{*}{101.55} & \multirow{2}{*}{2.03} \\
            & Random  & 3.98 & 1.51 &                         &                      \\
        \midrule
        \multirow{2}{*}{KDD}
            & Matched & 1.95 & 1.40 & \multirow{2}{*}{125.69} & \multirow{2}{*}{2.51} \\
            & Random  & 6.32 & 2.02 &                         &                      \\
        \bottomrule
    \end{tabular}
    }
\end{table}

\textbf{Practical Considerations in High Dimensions}. While Euclidean distances can become less informative in extremely high-dimensional sparse
spaces, the observed large effect sizes indicate that the proposed matching remains
discriminative for the datasets considered. For more challenging regimes, we recommend
mitigating the curse of dimensionality by: (i) applying dimensionality reduction prior to
KNN matching to improve metric reliability, or (ii) substituting standard Optimal Transport
with Sliced Optimal Transport, which often provides improved statistical robustness and
computational efficiency in high-dimensional feature spaces.

\section{Computational Complexity of FairSHAP}
\label{app:complexity}

\textbf{Setup.} The dataset is split by the sensitive attribute into \(\mathcal{G}\in\mathbb{R}^{n\times d}\) and \(\widetilde{\mathcal{G}}\in\mathbb{R}^{m\times d}\) with \(d\) features. Let \(C_f\) be the cost of one forward pass of the model \(f\). The permutation Shapley estimator uses \(M\) permutations per instance and \(R\) reference draws per set‑function evaluation. With entropic optimal transport (OT) matching, let \(I\) denote the number of Sinkhorn iterations. A \emph{one‑side} run means a single call to \texttt{FairSHAP} with a fixed target group and the other as reference (e.g., target \(=\mathcal{G}\), reference \(=\widetilde{\mathcal{G}}\)). The \emph{both‑sides} setting refers to the full framework that runs \texttt{FairSHAP} twice with roles swapped and then concatenates the two modified groups.

\textbf{Matching.} With nearest‑neighbor (NN) matching, forming a dense cost matrix over all \(n\times m\) pairs in \(d\) dimensions costs \(O(nmd)\) time; normalizing and taking rowwise argmax adds \(O(nm)\). Storing the joint matrix \(\mathcal{P}\) uses \(O(nm)\) space. With entropic OT, the total time is \(O(nmd+Inm)\) and the space remains \(O(nm)\).

\textbf{Shapley estimation (one side).} Using a permutation prefix‑chain estimator, each of the \(n\) target instances performs \(M d\) set‑function evaluations; each is approximated with \(R\) reference draws and one \(\mathrm{DR}\) evaluation of cost \(\mathcal{O}(C_f)\). The time is
\[
T_{\text{Shapley}} = \underbrace{ \mathcal{O}\big(n\,M\,d\,R\,C_f\big)}_{\text{No more complex than standard SHAP}} 
\,, \qquad \text{space } \mathcal{O}(nd) \,.
\]
Sampling from \(\mathcal{P}(\cdot\mid \mathbf{g}_i)\) can be made \(\mathcal{O}(1)\) per draw after \(\mathcal{O}(m)\) preprocessing per row, which does not change the asymptotics dominated by model evaluations.

\paragraph{Reference construction and edits.}
After obtaining the joint matrix $\mathcal{P}$, each target instance $i$
selects a reference index $j^{*}$ either by
(i) deterministic row-wise $\arg\max_j \mathcal{P}_{i,j}$, which costs $\mathcal{O}(nm)$ overall,
or (ii) stochastic sampling of $R$ references per row, which costs $\mathcal{O}(nR)$ in total after an $\mathcal{O}(m)$ preprocessing per row.
Building the reference table $\mathcal{B}$ by copying the selected reference feature vectors requires $\mathcal{O}(nd)$ time and space.
Subsequently, thresholding the Shapley values and replacing the corresponding features in $\mathcal{B}$ adds another $\mathcal{O}(nd)$.
Thus the overall cost of this stage in (i) is
\[
T_{\mathrm{ref}} = \mathcal{O}(nm + nd) \,,
\qquad
S_{\mathrm{ref}} = \mathcal{O}(nd) \,,
\]
and can be streamed to avoid extra peak memory.

\paragraph{End-to-end cost (one side).}
With NN matching:
\[
T_{\mathrm{one}} = \mathcal{O}\big(nmd + nMdRC_f\big) \,,\qquad
S_{\mathrm{one}} = \mathcal{O}\big(nm + nd\big) \,.
\]

\paragraph{Overall framework (both sides).}
The dense pairwise cost/similarity matrix can be computed once and then reused (or transposed) when swapping roles, so the $O(nmd)$ term is incurred once, while Shapley scales with the number of target instances on each side:
\[
T_{\mathrm{total}} = \mathcal{O}\big(nmd + (n{+}m)\,MdRC_f\big) \,,\qquad
S_{\mathrm{total}} = \mathcal{O}\big(nm + (n{+}m)d\big) \,.
\]

\paragraph{Remarks.}
(i) When model inference is expensive, the Shapley term $\mathcal{O}((n{+}m)MdRC_f)$ dominates; for very large $n,m$ and moderate $d$, the $\mathcal{O}(nmd)$ matching can dominate.  
\\
(ii) Exact Shapley per side requires $\mathcal{O}(n\,2^{d}\,R\,C_f)$ time, and the computation grows exponentially with $d$, quickly becoming infeasible beyond small feature dimensions.
\\
(iii) Compared with existing alternatives, 
KernelSHAP requires $\mathcal{O}(M C_f + M d^2)$ time per instance, where $M$ is the number of sampled coalitions (Exact Shapley corresponds to the special case $M=2^d$, which is exponential in $d$). 
For tree ensembles, TreeSHAP achieves polynomial-time computation, with complexity $\mathcal{O}(T L D^2)$ in the number of trees $T$, maximum leaves $L$, and depth $D$ (often simplified to $\mathcal{O}(d^2)$ when $D\!\approx\! d$). 
Our prefix-chain estimator reduces the per-side cost to $\mathcal{O}(nMdRC_f)$. If more efficient Shapley value approximation methods become available, they can be plugged into the FairSHAP framework to further accelerate attribution without changing the overall structure.

\section{Threshold Selection}
\label{app:threshold}
\vspace{-1mm}
In this section, we justify the choice of the Shapley threshold $T=0.05$ used throughout our experiments. 
Recall that $T$ determines which features are considered fairness-critical and thus subject to modification: 
smaller thresholds yield fewer edits (potentially insufficient for fairness improvement), 
while larger thresholds increase the number of edits (at the risk of introducing noise and harming data fidelity).
\begin{figure}[h]
\centering
\includegraphics[width=1\linewidth]{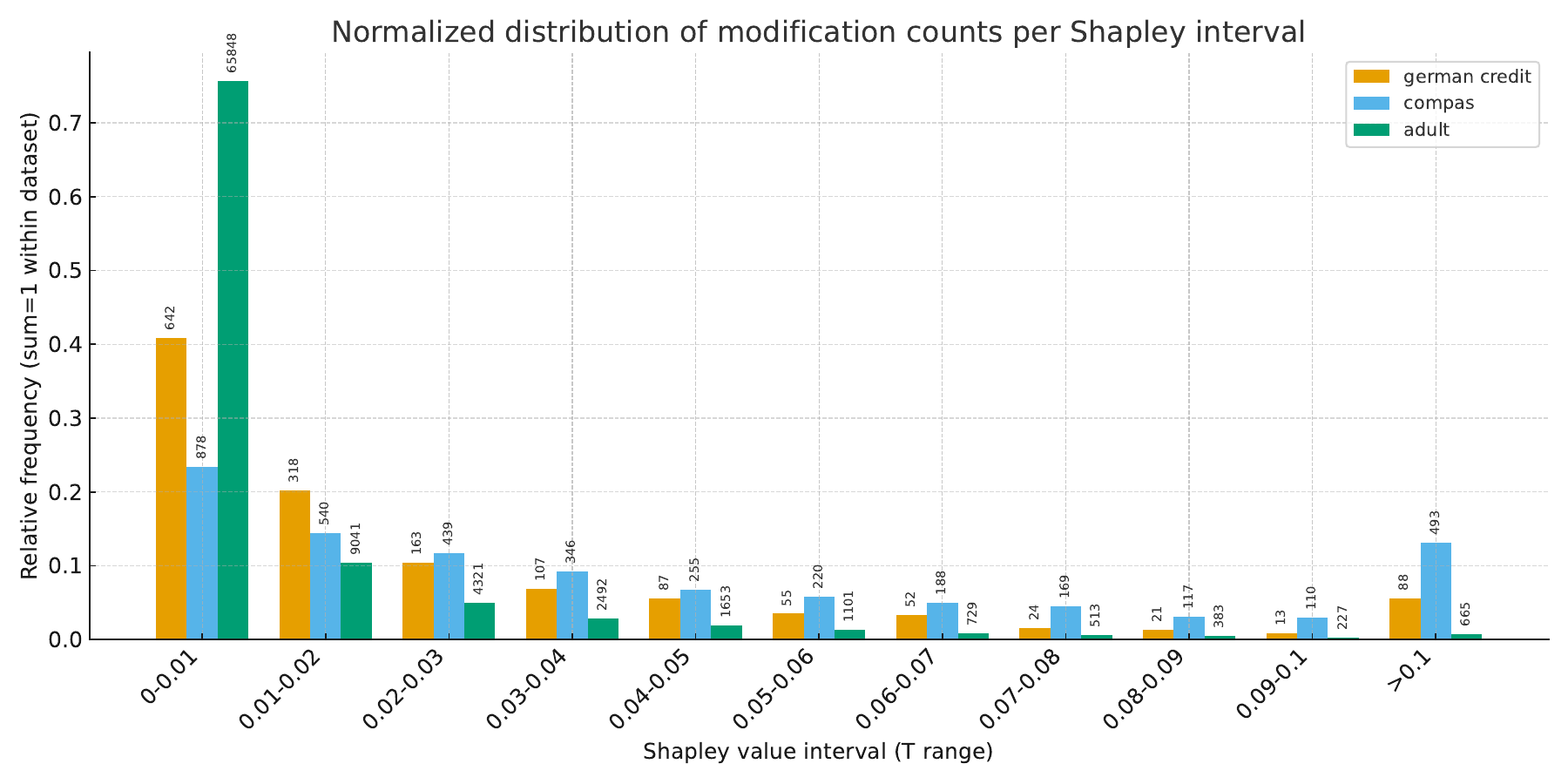}
\vspace{-7mm}
\caption{Normalized distribution of modification counts per Shapley interval. 
Each color corresponds to a dataset (German Credit, COMPAS, Adult). 
Y-axis shows relative frequency (summing to 1 within each dataset), 
while numbers above bars indicate the absolute counts of modifications in each interval.}
\label{fig:threshold_counts}
\end{figure}

\vspace{-4mm}
\subsection*{Distribution of modification counts}
To better understand the effect of thresholding, we analyze the distribution of feature modifications across different 
Shapley value intervals. For each dataset, we group the absolute Shapley values into bins of size 0.01 up to 0.10, 
and a final bin for values larger than 0.10. 
Figure~\ref{fig:threshold_counts} shows the normalized histograms: 
the bar heights are relative frequencies (summing to one within each dataset), 
while the numbers annotated above the bars indicate the raw modification counts.

\vspace{-2mm}
\subsection*{Observations}
\vspace{-1mm}
From below figure~\ref{fig:threshold_counts}, several patterns emerge:
\begin{itemize}[topsep=2pt,itemsep=1pt,parsep=0pt,partopsep=0pt]
  \item \textbf{Early growth and plateau.} Most edits concentrate within $|\phi|\in[0.01,0.05]$. 
  Beyond this range, additional intervals contribute relatively few new edits.
  \item \textbf{Stable mid-range.} Across datasets, the distributions flatten around $T=0.05$, 
  indicating that this value lies in a regime where edits remain meaningful but not excessive.
  \item \textbf{Avoiding extremes.} Very large thresholds ($>0.07$)
  fail to capture sufficient fairness-critical features, 
  while very small thresholds ($<0.03$) incorporate many near-zero contributions, 
  which may introduce noise and cost much time.
\end{itemize}

\vspace{-2mm}
\subsection*{Choice of $T=0.05$}
\vspace{-1mm}
Empirical results across datasets confirm that $T$ values in the range $[0.03,0.07]$ 
strike a favorable balance between fairness gains and data fidelity. 
We therefore adopt $T=0.05$ as a default threshold: 
\begin{enumerate}[topsep=2pt,itemsep=1pt,parsep=0pt,partopsep=0pt]
  \item it captures the majority of fairness-critical contributions without excessive edits;
  \item it avoids the ambiguity of near-zero Shapley values;
  \item it yields consistent improvements in both individual and group fairness metrics (DR, DP, EO), 
  with minimal accuracy degradation.
\end{enumerate}
This threshold thus represents a principled and empirically validated trade-off point 
for FairSHAP across diverse datasets. \textit{We further suggest that practitioners may tune the threshold on a validation set 
to identify the most suitable value for a given dataset and application.}

\section{Robustness across different ML models}
We acknowledge that Shapley values are inherently dependent on the underlying predictive model. 
Nevertheless, FairSHAP can naturally adapt to different architectures by identifying and mitigating those feature contributions that most strongly affect fairness for a given predictor. 
To evaluate the stability of our method across classifiers, we extend our experiments on the German Credit and COMPAS datasets using three representative models: XGBoost~\citep{chen2016xgboost}, SVM~\citep{cortes1995support}, and TabPFN~\citep{hollmann2025tabpfn}. 
The results in Table~\ref{tab:model_comparison} demonstrate that FairSHAP consistently improves fairness metrics across different model architectures, confirming its robustness to variations in predictive structure.
\begin{table}[h]
\centering\vspace{-3mm}
\caption{Model performance on German Credit and COMPAS datasets (mean $\pm$ std), 
with sex as the sensitive attribute. SHAP values are computed using KernelSHAP. 
FairSHAP results that outperform the baseline are highlighted in \textbf{bold}.}
\label{tab:model_comparison}\vspace{1mm}
\begin{adjustbox}{width=.937\textwidth, center}
\begin{tabular}{llc rrrr}
\toprule
\textbf{Model} & \textbf{Method} & \textbf{Accuracy} & \multicolumn{1}{c}{\textbf{DR}} & \multicolumn{1}{c}{\textbf{DP}} & \multicolumn{1}{c}{\textbf{EO}} & \multicolumn{1}{c}{\textbf{PQP}} \\
\midrule
\multicolumn{7}{c}{\textbf{German Credit (sex)}} \\
\cmidrule(r){1-7}
XGBoost & Baseline & 0.6650\topelement{0.0257} & 0.0785\topelement{0.0211} & 0.0512\topelement{0.0346} & 0.1287\topelement{0.0590} & 0.1341\topelement{0.0486} \\
        & FairSHAP & 0.6630\topelement{0.0275} & \textbf{0.0243\topelement{0.0112}} & \textbf{0.0301\topelement{0.0347}} & \textbf{0.1126\topelement{0.0783}} & 0.1852\topelement{0.1074} \\
\addlinespace
SVM     & Baseline & 0.7170\topelement{0.0175} & 0.0368\topelement{0.0139} & 0.0374\topelement{0.0327} & 0.0929\topelement{0.0948} & 0.3728\topelement{0.1957} \\
        & FairSHAP & 0.7100\topelement{0.0240} & \textbf{0.0306\topelement{0.0099}} & \textbf{0.0335\topelement{0.0643}} & 0.1138\topelement{0.1049} & \textbf{0.3185\topelement{0.0885}} \\
\addlinespace
TabPFN  & Baseline & 0.7170\topelement{0.0236} & 0.0354\topelement{0.0118} & 0.0609\topelement{0.0292} & 0.1429\topelement{0.1317} & 0.2863\topelement{0.1158} \\
        & FairSHAP & 0.7130\topelement{0.0271} & \textbf{0.0220\topelement{0.0057}} & \textbf{0.0477\topelement{0.0628}} & \textbf{0.1010\topelement{0.1204}} & 0.3027\topelement{0.1938} \\
\midrule
\multicolumn{7}{c}{\textbf{COMPAS (sex)}} \\
\cmidrule(r){1-7}
XGBoost & Baseline & 0.6698\topelement{0.0051} & 0.0883\topelement{0.0064} & 0.1548\topelement{0.0241} & 0.1243\topelement{0.0510} & 0.0492\topelement{0.0084} \\
        & FairSHAP & 0.6609\topelement{0.0106} & \textbf{0.0629\topelement{0.0091}} & \textbf{0.1326\topelement{0.0407}} & \textbf{0.0985\topelement{0.0603}} & \textbf{0.0452\topelement{0.0383}} \\
\addlinespace
SVM     & Baseline & 0.6759\topelement{0.0101} & 0.0284\topelement{0.0039} & 0.1688\topelement{0.0096} & 0.1717\topelement{0.0378} & 0.0658\topelement{0.0246} \\
        & FairSHAP & 0.6738\topelement{0.0129} & \textbf{0.0263\topelement{0.0056}} & \textbf{0.1542\topelement{0.076}}  & \textbf{0.1557\topelement{0.0359}} & \textbf{0.0598\topelement{0.0326}} \\
\addlinespace
TabPFN  & Baseline & 0.6876\topelement{0.0103} & 0.0548\topelement{0.0092} & 0.1981\topelement{0.0289} & 0.1771\topelement{0.0500} & 0.0234\topelement{0.0030} \\
        & FairSHAP & 0.6842\topelement{0.0122} & \textbf{0.0465\topelement{0.0044}} & \textbf{0.1614\topelement{0.0225}} & \textbf{0.1416\topelement{0.0527}} & 0.0458\topelement{0.0162} \\
\bottomrule
\end{tabular}
\end{adjustbox}
\end{table}

\section{Robustness across different SHAP kernels}
\label{app:appendix_shap_experiments}
We further examined robustness under different Shapley-value approximations. Specifically, we compared KernelSHAP~\citep{roshan2022using}, SamplingSHAP~\citep{lundberg2017unified}, and PermutationSHAP~\citep{strumbelj2010efficient} on German Credit and COMPAS with XGBoost fixed as the base model. As shown in Table~\ref{tab:robustness}, FairSHAP yields consistent fairness gains regardless of explainer, with KernelSHAP offering the best trade-off between accuracy, fairness, and runtime. These results indicate that FairSHAP’s effectiveness is not tied to a particular estimator and can readily benefit from future advances in scalable Shapley-value computation.
\begin{table}[h]
\centering
\caption{Comparison of FairSHAP performance using different Shapley value estimators on the German Credit and COMPAS datasets (mean $\pm$ std). 
All results are computed based on XGBoost, with sex as the sensitive attribute. 
Shapley values are approximated via KernelSHAP, SamplingSHAP, and PermutationSHAP, respectively.}
\label{tab:robustness}\vspace{1mm}
\begin{adjustbox}{width=.9367\textwidth, center}
\begin{tabular}{lccccc r}
\toprule
\textbf{Method} & \textbf{Accuracy} & \textbf{DR} & \textbf{DP} & \textbf{EO} & \textbf{PQP} & \textbf{Time (s)} \\
\midrule
\multicolumn{7}{c}{\textbf{German Credit (sex)}} \\
\cmidrule(r){1-7}
Baseline                 & 0.665\topelement{0.026} & 0.079\topelement{0.021} & 0.051\topelement{0.035} & 0.129\topelement{0.059} & 0.134\topelement{0.049} & --- \\
Kernel explainer         & 0.663\topelement{0.028} & 0.024\topelement{0.011} & 0.030\topelement{0.035} & 0.113\topelement{0.078} & 0.185\topelement{0.107} & 19 \\
Sampling explainer       & 0.668\topelement{0.039} & 0.027\topelement{0.011} & 0.043\topelement{0.036} & 0.169\topelement{0.086} & 0.178\topelement{0.137} & 224 \\
Permutation explainer    & 0.660\topelement{0.027} & 0.024\topelement{0.014} & 0.043\topelement{0.031} & 0.118\topelement{0.100} & 0.195\topelement{0.103} & 42 \\
\midrule
\multicolumn{7}{c}{\textbf{COMPAS (sex)}} \\
\cmidrule(r){1-7}
Baseline                 & 0.670\topelement{0.005} & 0.088\topelement{0.006} & 0.155\topelement{0.024} & 0.124\topelement{0.051} & 0.049\topelement{0.008} & --- \\
Kernel explainer         & 0.661\topelement{0.011} & 0.063\topelement{0.009} & 0.133\topelement{0.041} & 0.099\topelement{0.060} & 0.045\topelement{0.038} & 40 \\
Sampling explainer       & 0.654\topelement{0.012} & 0.062\topelement{0.006} & 0.142\topelement{0.025} & 0.114\topelement{0.041} & 0.046\topelement{0.039} & 556 \\
Permutation explainer    & 0.661\topelement{0.014} & 0.060\topelement{0.003} & 0.141\topelement{0.027} & 0.115\topelement{0.044} & 0.046\topelement{0.040} & 382 \\
\bottomrule
\end{tabular}
\end{adjustbox}
\end{table}

\clearpage
%

\clearpage
\appendix
\thispagestyle{empty}
%
\end{document}